\documentclass[letterpaper]{article} 
\usepackage{aaai24}  
\usepackage{times}  
\usepackage{helvet}  
\usepackage{courier}  
\usepackage[hyphens]{url}  
\usepackage{graphicx} 
\usepackage{natbib}  
\usepackage{caption} 
\usepackage{algorithm}
\usepackage{algorithmic}

\usepackage{newfloat}
\usepackage{listings}
\DeclareCaptionStyle{ruled}{labelfont=normalfont,labelsep=colon,strut=off} 
\floatstyle{ruled}
\newfloat{listing}{tb}{lst}{}
\floatname{listing}{Listing}
\usepackage{amsfonts,mathtools}
\usepackage{booktabs,utfsym}
\usepackage{subcaption}
\usepackage{xfrac}
\usepackage{multirow,multicol}
\title{What Makes Good Collaborative Views? Contrastive Mutual Information Maximization for Multi-Agent Perception}
\author{
    Wanfang Su\equalcontrib \textsuperscript{\rm,1,2},
    Lixing Chen\equalcontrib \textsuperscript{\rm,1,2},
    Yang Bai\thanks{Corresponding authors.}\textsuperscript{\rm,1},
    Xi Lin\textsuperscript{\rm 1,2},
    Gaolei Li\textsuperscript{\rm 1,2},
    Zhe Qu\textsuperscript{\rm 3},
    Pan Zhou\textsuperscript{\rm 4}
}
\affiliations{
    \textsuperscript{\rm 1}School of Electronic Information and Electrical Engineering, Shanghai Jiao Tong University, Shanghai, China\\
    \textsuperscript{\rm 2}Shanghai Key Laboratory of Integrated Administration Technologies for Information Security, Shanghai, China\\
    \textsuperscript{\rm 3}School of Computer Science and Engineering, Central South University, Changsha, China\\
    \textsuperscript{\rm 4} Hubei Engineering Research Center on Big Data Security, School of Cyber Science and Engineering, Huazhong University of Science of Technology, Wuhan, China

    \{77susu, lxchen, ybai, linxi234, gaolei\_li\}@sjtu.edu.cn, zhe\_qu@csu.edu.cn, zhoupannewton@gmail.com
}

\usepackage{bibentry}

\usepackage{soul}
\ifodd 1

\else

\fi

\begin{document}
\maketitle
\begin{abstract}
Multi-agent perception (MAP) allows autonomous systems to understand complex environments by interpreting data from multiple sources. This paper investigates intermediate collaboration for MAP with a specific focus on exploring ``good'' properties of collaborative view (i.e., post-collaboration feature) and its underlying relationship to individual views (i.e., pre-collaboration features), which were treated as an opaque procedure by most existing works. We propose a novel framework named CMiMC (\underbar{C}ontrastive \underbar{M}utual \underbar{I}nformation \underbar{M}aximization for \underbar{C}ollaborative Perception) for intermediate collaboration. The core philosophy of CMiMC is to preserve discriminative information of individual views in the collaborative view by maximizing mutual information between pre- and post-collaboration features while enhancing the efficacy of collaborative views by minimizing the loss function of downstream tasks. In particular, we define multi-view mutual information (MVMI) for intermediate collaboration that evaluates correlations between collaborative views and individual views on both global and local scales. We establish CMiMNet based on multi-view contrastive learning to realize estimation and maximization of MVMI, which assists the training of a collaboration encoder for voxel-level feature fusion. We evaluate CMiMC on V2X-Sim 1.0, and it improves the SOTA average precision by 3.08\% and 4.44\% at 0.5 and 0.7 IoU (Intersection-over-Union) thresholds, respectively. In addition, CMiMC can reduce communication volume to $\sfrac{1}{32}$ while achieving performance comparable to SOTA. Code and Appendix are released at https://github.com/77SWF/CMiMC.  
\end{abstract}

\section{Introduction}
Multi-Agent perception (MAP) \citep{wang2023core,xu2023bridging} refers to the process by which multiple entities work together to gather a interpret sensory information and establish a collective comprehension of the environment. MAP enhances perception capabilities over the single-agent perception by mitigating inherent limitations arising from individual perspectives \citep{li2021learning}. For example, using MAP in autonomous driving \citep{cui2022coopernaut,chen2017multi} effectively addresses challenges caused by occlusions and sparse observations in long-distance areas. 

Collaboration strategy is one of the fundamental components of MAP, which involves protocols for communication, data sharing, and coordination among multiple agents. State-of-the-art collaboration strategies can be categorized into three types: early collaboration, late collaboration, and intermediate collaboration. \emph{Early Collaboration} \cite{chen2019cooper,arnold2020cooperative} aggregates the raw measurement from agents to generate a comprehensive view. However, it requires substantial communication bandwidth and incurs privacy leakage risk. \emph{Late Collaboration} \cite{miller2020cooperative} aggregates perception results of individual agents, which is bandwidth-efficient. However, individual perception results are noisy and coarse-grained, and therefore may cause inferior performance for collaborative perception. To strike a balance between performance and bandwidth usage, \emph{Intermediate Collaboration} has been proposed \citep{liu2020who2com, hu2022where2comm} to aggregate intermediate features of agents. The intermediate features distill compact knowledge representations from raw measurement, offering the potential for both communication efficiency and enhanced perception capability. Nevertheless, intermediate collaboration strategies require judicious design, otherwise, may result in substantial information loss and adversely impact the effectiveness of MAP. 

Recent endeavors have been dedicated to the advancement of intermediate collaboration strategies. For example, V2VNet \citep{wang2020v2vnet} proposes to exchange intermediate features and integrate them via a spatial-aware graph neural network for improving motion prediction performance; When2com \citep{liu2020when2com} constructs a communication group based on general attention mechanisms to determine the optimal timing for intermediate collaboration; DiscoNet \citep{li2021learning} trains a distilled collaboration network to push the results of intermediate collaboration close to that of early collaboration. The majority of current works treat the intermediate collaboration as an opaque process and approximate it with deep neural networks that are trained to optimize the performance of downstream tasks or other specified design goals. However, such a process can easily encounter performance bottlenecks as it runs the risk of losing valuable features solely in the pursuit of optimizing the pre-defined objectives. Evidence in \citet{hjelm2018learning} demonstrates that features generated by unsupervised learning outperform those generated by supervised learning that minimizes downstream loss. This paper aims to delve into fundamental principles of feature aggregation in intermediate collaboration, exploring the characteristics that contribute to good collaborative views (i.e., post-collaboration features) and how good collaborative views relate to views of individual agents (i.e., pre-collaboration features). 

Intuitively, it is desirable for collaborative views to aggregate all discriminative information in individual views. Drawing upon this intuition, we introduce the concept of mutual information (MI) maximization into intermediate collaboration. MI maximization is initially used in representation learning \citep{hjelm2018learning,linsker1988self} to find features that capture the underlying dependencies in the data. We extend its ability to intermediate collaboration for obtaining collaborative views that retain informative parts of individual views and discard irrelevant or redundant ones. The \textbf{contributions} of this paper are summarized as follows:

1. We propose a novel framework named CMiMC (\underbar{C}ontrastive \underbar{M}utual \underbar{I}nformation \underbar{M}aximization for \underbar{C}olloabrative perception) for intermediate collaboration. CMiMC defines multi-view mutual information (MVMI) that properly measures the global and local dependencies between a collaborative view and multiple individual views. We design an MVMI maximization strategy and plug it into the supervised learning process of MAP. This enables CMiMC to construct collaborative views that retain discriminative information of individual views and also provide enhanced effectiveness to downstream tasks.  

2. We establish CMiMNet based on multi-view contrastive learning to estimate and maximize MVMI in an unsupervised fashion by drawing close the collaborative view and individual views that come from the same scene (i.e., positive pairs) and pushing away individual views from different scenes (i.e., negative pairs). CMiMNet enables the intermediate collaboration strategy to identify critical regions in individual views and complete fine-grained feature aggregation at the voxel resolution.

3. We evaluate our method on V2X-Sim 1.0 \citep{li2021learning}. Experimental results show that CMiMC outperforms state-of-the-art (SOTA) benchmarks in terms of average precision (AP) and performance-bandwidth trade-offs. CMiMC improves SOTA AP by 3.08\%$\backslash$4.44\% at Intersection-over-Union (IoU) $0.5\backslash0.7$. It can reduce communication volume to \sfrac{1}{32} while achieving AP comparable to SOTA. 

\section{Related Work}
\label{sec:related}
\subsubsection{Collaborative Perception}
The existing works of collaborative strategies can be categorized into three types: Early Collaboration \citep{chen2019cooper} involves sharing raw measurement, Late Collaboration \citep{miller2020cooperative} involves sharing perception results, and Intermediate Collaboration involves sharing intermediate features. Among these approaches, intermediate collaboration has attracted significant attention as it strikes a balance between perception performance and communication bandwidth. For example, 
When2com \citep{liu2020when2com} forms a communication group based on the general attention mechanism to determine the optimal timing for intermediate collaboration. V2VNet~\citep{wang2020v2vnet} broadcasts and receives intermediate features using a spatially aware GNN to improve motion prediction performance. DiscoNet~\citep{li2021learning} utilizes knowledge distillation to push the results of intermediate collaboration close to that of early collaboration. Where2comm~\citep{hu2022where2comm} proposes a spatial confidence map to reflect the spatial heterogeneity of perceptual information. V2X-ViT~\citep{xu2022v2x} fuses information across heterogeneous agents using a multi-agent attention module. CRCNet~\citep{luo2022complementarity} studies feature selection by sequentially incorporating features that differ the most from fused ones. Most of the above works treat feature aggregation of intermediate collaboration as a black box and directly use deep neural networks to learn collaboration strategies that optimize downstream tasks or other specified goals. By contrast, our work focuses on the feature aggregation stage, aiming to unravel the characteristics that contribute to the formation of a good collaborative view. 

\subsubsection{Mutual Information Estimation and Maximization}
Mutual information (MI) is commonly used for quantifying the correlation between random variables. This paper leverages MI maximization to enhance the performance of intermediate collaboration. Existing works have studied MI estimation and maximization. Early non-parametric methods include binning~\citep{darbellay1999estimation}, kernel density estimator~\citep{silverman1981using}, and density-ratio-based likelihood estimator~\citep{hido2011statistical}. However, these works are limited to handling low-dimensional data. Recent works introduce neural networks into MI estimation, which has proven effective for handling high-dimensional data. For example, MINE~\citep{belghazi2018mutual} and InfoNCE \citep{oord2018representation} estimate MI by learning to maximize variational bounds. Deep InfoMax~\citep{hjelm2018learning}, Info3D~\citep{sanghi2020info3d} 
studies unsupervised representation learning by maximizing MI between encoder input and output. However, these works only consider the MI estimation for two variables. To address this issue, Contrastive Multiview Coding (CMC)~\citep{tian2020contrastive} is proposed to capture dependencies among multiple views. Inspired by these works, we introduce MI maximization to intermediate collaboration for constructing collaborative views that retain discriminative information of individual views. We designed multi-view mutual information (MVMI) that is tailored for intermediate collaboration to measure the dependencies of multiple variables in a one-to-many setting (i.e., one collaborative view to many individual views). MVMI also takes into account MI evaluations on both global and local scales. In addition, we design CMiMNet to estimate and maximize MVMI based on contrastive learning.

\section{Methods} \label{sec:methods}
For ease of presentation, we introduce CMiMC in the context of LiDAR-based 3D object detection. However, we note that CMiMC is also compatible with other MAP scenarios.

\begin{figure*}[tb]
\centering
\includegraphics[width= 1\linewidth]{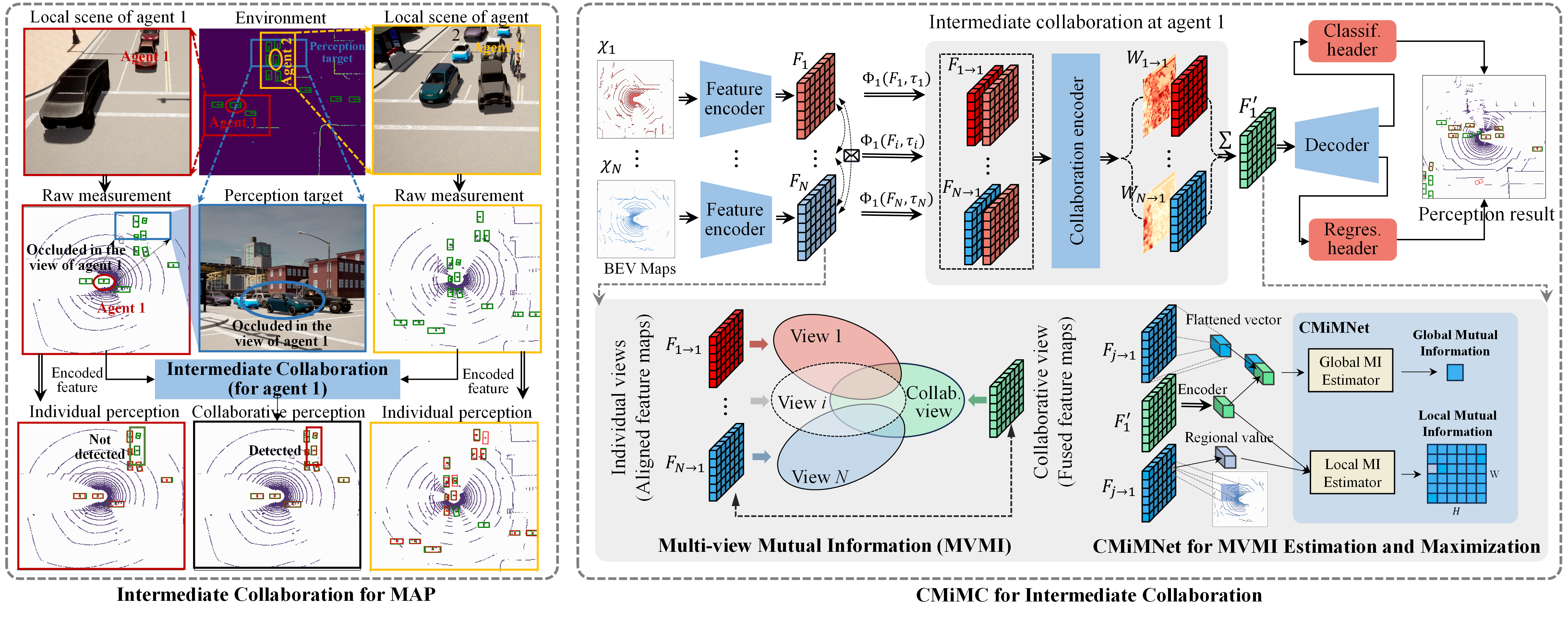}
\caption{Scenario of intermediate collaboration for LiDAR-based 3D object detection and the framework of CMiMC.}
\label{fig_framework}
\end{figure*}

\subsection{The CMiMC Framework}
We illustrate the scenario of intermediate collaboration for LiDAR-based 3D object detection and the framework of CMiMC in Fig.~\ref{fig_framework}. The upper part of CMiMC shows the pipeline of MAP, which comprises three parts: 1) a feature encoder for extracting features from raw measurements; 2) a collaboration encoder for aggregating features received from other agents; 3) a decoder-and-header module for decoding fused features and generating perception results. The bottom part of CMiMC illustrates the MI maximization module which encompasses the definition of Multi-View Mutual Information (MVMI) on global/local scales and CMiMNet for estimating and maximizing MVMI. 

All agents utilize the same feature encoder, collaboration encoder, and decoder-header module to perform collaborative perception. CMiMNet is exclusively used during training, and agents do not need to execute CMiMNet during inference. In the following, we give the procedures for performing the above components. The architectures of encoders/decoders are given in Appendix A.1.  

\subsubsection{Feature Encoder}
Each agent possesses a feature encoder to extract features from raw measurements. In the context of LiDAR-based 3D object detection, the raw measurements are 3D point clouds which will be converted into Bird's-Eye View (BEV) maps (see Appendix A.2 for details) and fed to the feature encoder. The BEV map of agent $i$ is denoted by $\chi_i$. The feature map extracted by the feature encoder is $F_i=\Psi_\texttt{enc}(\chi_i) \in \mathbb{R}^{H\times W\times C}$, where $\Psi_\texttt{enc}$ denotes the feature encoder and $H, W$, and $C$ denote the height, width, and channel of the intermediate feature map. We also call the feature map $F_i$ the \emph{individual view} of agent $i$. 

\subsubsection{Collaboration Encoder}
Agents exchange individual views via peer-to-peer communication and utilize a collaboration encoder to generate a collaborative view (i.e., a fused feature map). Because each agent perceives the environment using a local coordinate system with itself as the origin, an agent needs to align the received views before feature aggregation. We consider that agents exchange pose information along with intermediate features, and let $\tau_i$ denote the pose information of agent $i$. The feature alignment at agent $i$ for feature $F_j$ with pose information $\tau_j$ is denoted as $F_{j \rightarrow i} = \Phi_i(F_j,\tau_j)$, where $\Phi_i$ is a feature pose transformation function based on pose information $\tau_i$. Given the aligned feature maps, the collaboration encoder outputs spatial weight matrices for voxel-level feature aggregation. The weight matrix for aggregating agent $j$'s feature at agent $i$ is $W_{j \rightarrow i} = \Psi_\texttt{col}(F_{j \rightarrow i}, F_i) $, where $\Psi_\texttt{col}$ is the collaboration encoder. The weight matrix $W_{j \rightarrow i}$ has the same size as $F_{j \rightarrow i}$. Each element in $W_{j \rightarrow i}$ corresponds to a voxel at the same position in $F_{j \rightarrow i}$ and reflects the importance of the regional feature. Finally, agent $i$ can obtain a fused feature map by
\begin{equation}       
\label{equ_fuse}
F'_i = \sum\nolimits_{j=1}^{N} \Psi_\texttt{col}(F_{j \rightarrow i}, F_i) \otimes \Phi_i (F_j,\tau_j),
\end{equation}
where $N$ is the number of agents in the multi-agent perception task and $\otimes$ denotes element-wise multiplication.

\subsubsection{Decoder-Header Module}
Given collaborative view $F'_i$, the decoder-header module first decodes it to a feature map $\varrho_i = \Psi_\texttt{dec}(F'_i)$ via decoder $\Psi_\texttt{dec}$. The decoded feature is then fed to a classification header (for classifying the foreground-background categories) and a regression header (for generating the bounding boxes). The perception results at agent $i$ can be represented by $Y_i = \Psi_\texttt{head}(\varrho_i)$, where $\Psi_\texttt{head}$ denotes a combination of headers. 

\subsubsection{Multi-view Mutual Information Maximization for Intermediate Collaboration} 
\label{subsec_MiMax_module}
Agents in MAP capture diverse views of the environment, each offering a unique insight. Empirical evidence in numerous real-world scenarios \citep{tian2020contrastive} supports the conventional belief that valuable information often resides in different views. However, how to gather and consolidate valuable information dispersed across individual views with minimal information loss is still challenging. To address this challenge, we introduce mutual information maximization into intermediate collaboration. Unlike previous approaches that construct collaborative views by minimizing the loss function of downstream tasks, our method aims to retain the discriminative features of individual views in the collaborative view by maximizing mutual information between them. 

We use mutual information (MI) to quantify the dependencies between individual views and collaborative views. Consider an individual view $F \in \{F_{1 \rightarrow i}, \dots, F_{N\rightarrow i}\}^N_{i=1}$ and a collaborative view $F^\prime \in \{F_i^\prime\}^N_{i=1}$, the MI between $F$ and $F^\prime$ is defined by $\mathcal{I}(F,F^\prime) = \int_{F \times F^\prime} \log \frac{{\rm d} \mathbb{P}_{FF^\prime}}{{\rm d} \mathbb{P}_{F}\mathbb{P}_{F^\prime}} {\rm d} \mathbb{P}_{FF^\prime}$. The Kullback-Leibler (KL) divergence between $\mathbb{P}_{FF^\prime}$ and $\mathbb{P}_{F} \mathbb{P}_{F^\prime}$ equals the above MI definition ~\citep{belghazi2018mutual}, and hence we can use KL divergence to reflect the overall dependency between an individual view and a collaborative view, which we call global MI: $\mathcal{I}_\texttt{G} (F, F^\prime) = D_{\texttt{KL}}(\mathbb{P}_{FF^\prime}||\mathbb{P}_{F} \mathbb{P}_{F^\prime})$. 
Global MI has been widely used in representation learning, however, it also has been shown to be insufficient for perception tasks when the structure of features matter~\cite{bachman2019learning}. Therefore, we further use local MI to evaluate the dependencies between the collaborative view and regional feature patches of the individual views. By doing so, we can properly capture the spatial structure of individual views. The local MI can be obtained by:
\begin{equation}
\label{equ_local_MI}
\small
\mathcal{I}_\texttt{L} (F,F') = \frac{1}{H\times W} \sum\nolimits_{k=1}^{H\times W} \mathcal{I}(F(k),F'),
\end{equation}
where $H$ and $W$ are the height and width of $F$ and $F(k)$ denote $k$-th regional feature patch in $F$. 

Vanilla MI is designed to measure the dependency between two random variables and does not match the scenario of intermediate collaboration where an agent needs to measure MI between one collaborative view and multiple individual views. To fill this gap, we define Multi-View Mutual Information (MVMI) which is tailored to this one-to-many MI evaluation setting. 

To be specific, the collaborative view $F^\prime_i$ is the core view for the collaboration encoder to optimize. MVMI first builds pair-wise MI (both global MI and local MI) between $F^\prime_i$ and each individual view $F_{j \to i}$. Then, pair-wise MIs are averaged to form MVMI. Mathematically, MVMI for agent $i$ is calculated as:
\begin{equation}
\label{equ_MVMI}
\small
\mathcal{I}_{\texttt{MV},i} = \underbrace{\frac{\beta_\texttt{G}}{N}\sum_{j=1}^N \mathcal{I}_\texttt{G}(F_{j \to i},F_i^\prime)}_{\text{Global MVMI}} + \underbrace{\frac{\beta_\texttt{L}}{N}\sum_{j=1}^N \mathcal{I}_\texttt{L}(F_{j \to i},F_i^\prime)}_{\text{Local MVMI}},
\end{equation}
where $\beta_\texttt{G}$ and $\beta_\texttt{L}$ are weights for global and local MVMI. 
However, MI is notoriously difficult to compute as underlying distributions of features (i.e., $\mathbb{P}_{F}$, $\mathbb{P}_{F^\prime}$ and $\mathbb{P}_{FF^\prime}$) are unknown in practice, particularly for continuous and high-dimensional intermediate features in our work. In the following, we build CMiMNet based on contrastive learning to realize the estimation and maximization of MVMI.

\subsection{CMiMNet}
We now present CMiMNet for MVMI maximization.

\subsubsection{MVMI Maximization via Contrastive Learning}
\begin{figure}[b]
\centering
\includegraphics[width= 0.95\linewidth]{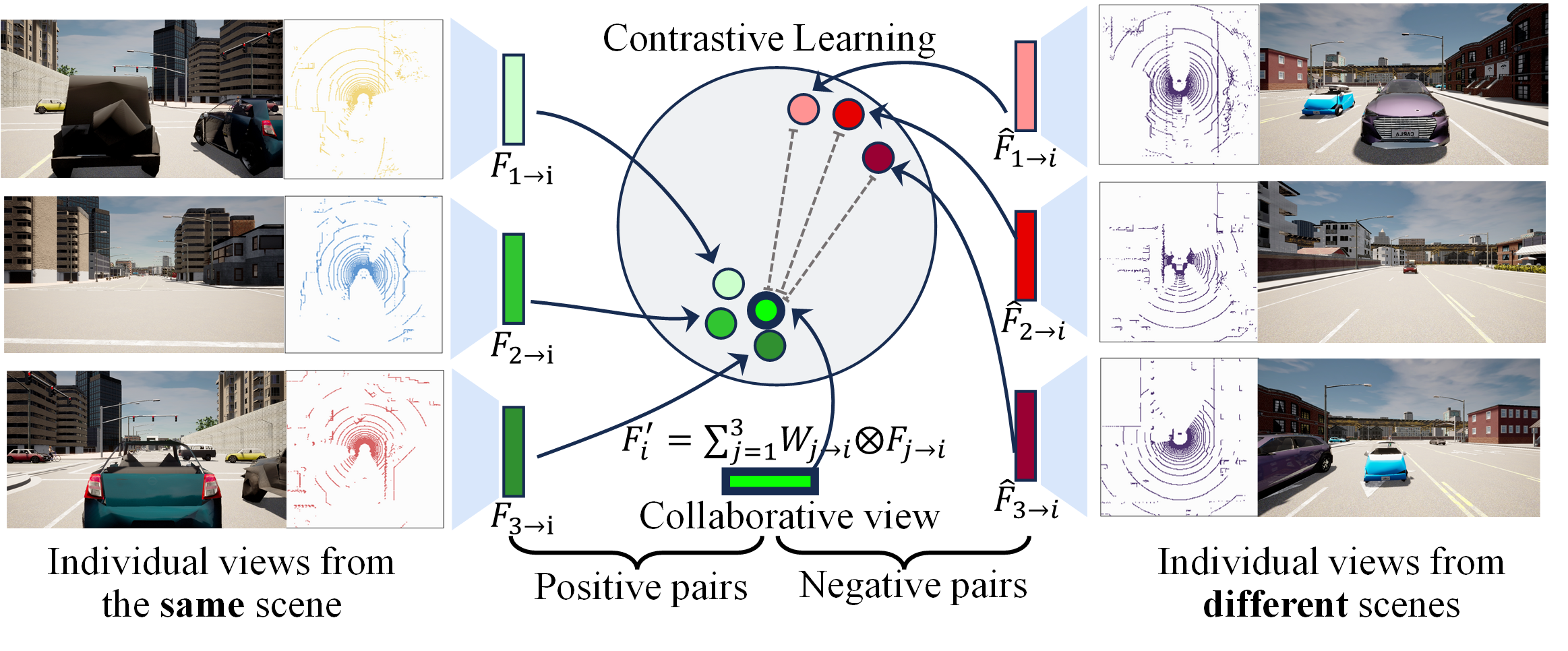}
\caption{Contrastive learning over positive/negative pairs.}
\label{fig_positive_negative_pairs}
\end{figure}

The goal of contrastive learning is to learn a discriminator $T_\theta(\cdot)$ (parameterized by $\theta$) to distinguish if two views, e.g., $F$ and $F^\prime$, come from different distributions $\mathbb{P}_{FF^\prime}$ and $\mathbb{P}_F \mathbb{P}_{F^\prime}$. The discriminator $T_\theta(\cdot)$ is trained to minimize the contrastive loss by assigning a high score to positive sample pair (i.e., samples from the same distribution) $(F_{j\rightarrow i}, F^\prime_i)_{\forall j\in [1,N]} \sim \mathbb{P}_{FF^\prime}$ and low score to negative sample pair (i.e., samples from different distributions) $(F_{ j \rightarrow k}, F'_i)_{\forall j\in [1, N], k \neq i} \sim \mathbb{P}_F \mathbb{P}_{F^\prime}$. To put it understandable in our context (Fig. \ref{fig_positive_negative_pairs}), $T_\theta$ draws close collaborative views and individual views that come from the same scene and pushes away individual views from different scenes. The loss of $T_\theta$ is termed as contrastive loss $\mathcal{L}_\texttt{CTRS}$, which can be used to estimate the lower bound of MI \citep{oord2018representation,poole2019variational}, i.e.,
\begin{equation}
\label{equ_MI_lowerbound}
\small
\mathcal{I}(F,F^\prime) \geq \log(K) - \mathcal{L}_\texttt{CTRS}(T_\theta(\cdot,\cdot)) = \hat{\mathcal{I}}(F,F^\prime),
\end{equation}
where $K$ represents the number of negative sample pairs.
Minimizing the contrastive loss equivalently maximizes the lower bound of MI $\mathcal{I}(\cdot)$. Therefore, we can define the negative value of estimated MI $\hat{\mathcal{I}}(\cdot)$ as the contrastive loss. 

\begin{figure*}[tb]
\centering
\includegraphics[width= 0.9 \linewidth]{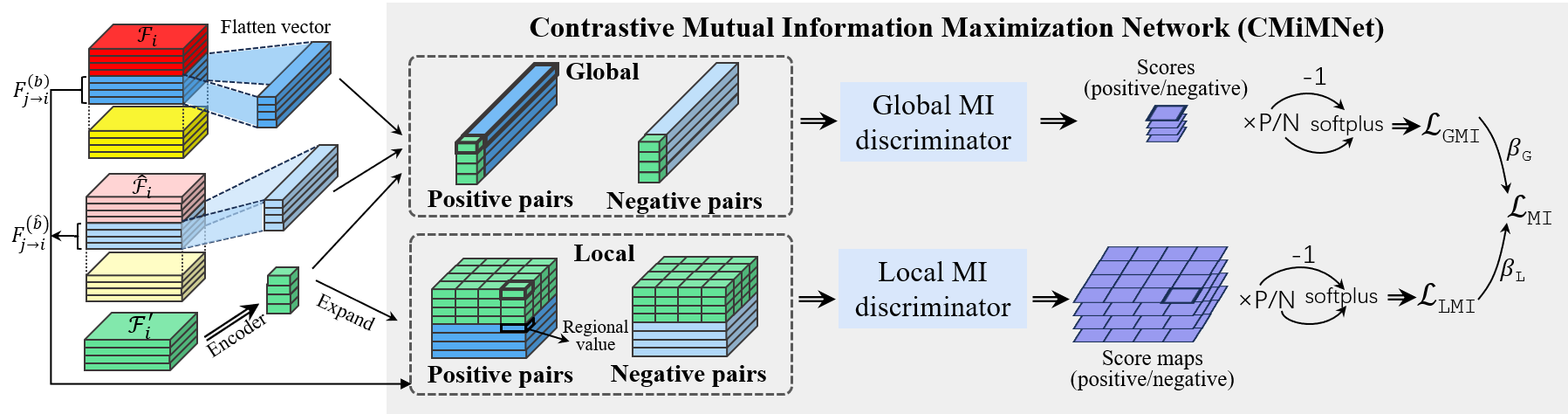}
\caption{The structure of CMiMNet for estimating and maximizing MVMI.}
\label{fig_cmimnet}
\end{figure*}

Now, the problem becomes estimating MI $\hat{\mathcal{I}}(\cdot)$. Existing works often use KL-divergence-based MI estimators, e.g., MINE \citep{belghazi2018mutual} and InfoNCE \citep{oord2018representation}, however, these methods require a large number of negative samples \citep{hjelm2018learning} to avoid unstable estimations. Therefore, CMiMNet utilizes an MI estimator based on Jensen-Shannon (JS) divergence \citep{nowozin2016f}, which delivers a similar capability as KL divergence-based MI estimators but exhibits higher stability \cite{hjelm2018learning}. The MI estimator is constructed on discriminator $T_\theta$, formally defined by $\hat{\mathcal{I}}(F,F^\prime;\theta) = \mathbb{E}_{\mathbb{P}_{FF^\prime}}[-\delta(-T_{\theta}(F, F^\prime))] -\mathbb{E}_{\mathbb{P}_F \mathbb{P}_{F^\prime}}[\delta(T_{\theta}(\hat F, F^\prime))]$, where $(F, F^\prime)$ and $(\hat F, F^\prime)$ are positive and negative sample pairs; $\delta(x) = \log(1+e^x)$ is the softplus function.

Based on MI estimator $\hat{\mathcal{I}}(\cdot)$ and definitions of global/local MI, we can write global MI estimator and local MI estimator as $\hat{\mathcal{I}}_\texttt{G}(F,F^\prime;\theta_\texttt{G}) = \hat{\mathcal{I}}(F,F^\prime;\theta_\texttt{G})$ and $\hat{\mathcal{I}}_{\texttt{L}}(F,F^\prime;\theta_\texttt{L}) = \frac{1}{H\times W}\sum_{k=1}^{H\times W} \hat{\mathcal{I}}(F(k),F^\prime;\theta_\texttt{L})$, where $\theta_\texttt{G}$ and $\theta_\texttt{L}$ are the parameters of global and local discriminators, respectively. Given the definition of MVMI in \eqref{equ_MVMI}, the objective of CMiMNet can be written as:
\begin{equation} \label{equ_obj_MMI}
\small
\max\limits_{\theta_\texttt{G},\theta_\texttt{L}} ~
 \frac{\beta_\texttt{G}}{N}\sum_{j=1}^N \hat{\mathcal{I}}_\texttt{G}(F_{ j \to i}, F^\prime_i;\theta_\texttt{G}) + \frac{\beta_\texttt{L}}{N}\sum^N_{j=1}\hat{\mathcal{I}}_\texttt{L}(F_{ j \to i}, F^\prime_i;\theta_\texttt{L}). 
\end{equation}

\subsubsection{Positive and Negative Pairs for CMiMNet}
CMiMNet learns in an unsupervised fashion using positive and negative sample pairs. In our problem, we pair collaborative views and individual views sampled from the same perception scene as positive pairs, and pair the collaborative views and individual views from different scenes as negative pairs (see Fig. \ref{fig_positive_negative_pairs} as an example). In each training iteration, we sample $B$ scenes, denoted by $\mathcal{B} = \{1,2,\dots,B\}$. For each agent $i$, we collect its collaborative views of $B$ scenes in set $\mathcal{F}^\prime_i = \{{F'_i}^{(b)}\}_{b\in\mathcal{B}}$, and then collect the individual views for generating these collaborative views in set $\mathcal{F}_i = \{F_{1\rightarrow i}^{(b)}, F_{2\rightarrow i}^{(b)}, \dots, F_{N\rightarrow i}^{(b)}\}_{b \in \mathcal{B}}$. We sample another $B$ scenes, denoted by $\hat{\mathcal{B}}$, which are different from that in $\mathcal{B}$, i.e., $\hat{\mathcal{B}} \cap \mathcal{B} = \emptyset$. We get individual views of $N$ agents about scenes in $\hat{\mathcal{B}}$, and collect them in $\hat{\mathcal{F}}_i = \{F_{1\rightarrow i}^{(\hat{b})}, F_{2\rightarrow i}^{(\hat{b})}, \dots, F_{N\rightarrow i}^{(\hat{b})}\}_{\hat{b} \in \hat{\mathcal{B}}}$. Then, for $b$-th scene in $\mathcal{B}$, we can construct $N$ positive pairs $\{(F_{j\rightarrow i}^{(b)}, {F^\prime_i}^{(b)})\}_{j=1}^N$ and $N$ negative pairs $\{(F_{j\rightarrow i}^{(\hat{b})},{F'_i}^{(b)})\}_{j=1}^N$ for training global and local discriminators, i.e., $T_{\theta_\texttt{G}}$ and $T_{\theta_\texttt{L}}$, and consequently maximizing the objective in \eqref{equ_obj_MMI}.

\subsubsection{Global and Local MVMI Estimator}
The structure of the global and local MVMI estimator is presented in Fig.~\ref{fig_cmimnet}. For each sample pair $(F, F^\prime)$, the global MVMI estimator first encodes the collaborative view $F^\prime$ into a feature vector via a linear layer. Individual view $F$ are flattened and concatenated to the feature vector, and then fed to the global discriminator $T_{\theta_\texttt{G}}$. The outputs of $T_{\theta_\texttt{G}}$ are scores for positive/negative sample pair, which can be used for global MVMI estimation. For the local MVMI estimator, the collaborative view $F^\prime$ is also converted to a feature vector using the encoder. Then, the generated feature vector is concatenated with regional values of individual features, which gives a set of vectors $\{ [F(k), \omega(F')]\}_{k=1}^{H\times W}$ (where $\omega(\cdot)$ denotes the encoder). These vectors are fed to the local MI discriminator $T_{\theta_\texttt{L}}$, generating a $H\times W$ score map. The score maps of all positive and negative pairs can be used to estimate local MVMI. The architectures of global and local discriminators are given in Appendix A.1.

\subsubsection{Loss Function}
During training, CMiMC updates the parameters of CMiMNet and other components in MAP simultaneously in a direction of minimizing overall loss $\mathcal{L}(\boldsymbol\Psi, \theta_\texttt{G},\theta_\texttt{L})$ (where $\boldsymbol \Psi = (\Psi_\texttt{enc},\Psi_\texttt{col},\Psi_\texttt{dec},\Psi_\texttt{head})$):
\begin{equation}
\label{equ_L_overall}
\small
\mathcal{L}(\boldsymbol\Psi,\theta_\texttt{G},\theta_\texttt{L}) = (1-\alpha) (\mathcal{L}_\texttt{CLS} + \mathcal{L}_\texttt{REG}) + \alpha \mathcal{L}_\texttt{MI},
\end{equation}
In Eqn. \eqref{equ_L_overall}, $\mathcal{L}_\texttt{CLS}$ and $\mathcal{L}_\texttt{REG}$ are the losses for foreground-background classification and bounding box regression. They are associated with downstream tasks (i.e., LiDAR-based 3D object detection). $\mathcal{L}_\texttt{MI}$ is the loss for MVMI estimation. The variable $\alpha \in [0,1]$ is used to adjust the importance of loss terms. As we consider both global MVMI and local MVMI,  $\mathcal{L}_\texttt{MI}$ can be feather decomposed into $\mathcal{L}_\texttt{MI} = \lambda (\beta_\texttt{G} \mathcal{L}_\texttt{GMI} + \beta_\texttt{L} \mathcal{L}_\texttt{LMI})$, where $\lambda$ is the weight to re-scale the MI loss.
Based on the objective of CMiMNet in \eqref{equ_obj_MMI}, we can define $\mathcal{L}_\texttt{GMI}$ and $\mathcal{L}_\texttt{LMI}$ as $\mathcal{L}_\texttt{GMI} = -\frac{1}{N}\sum_{j=1}^N \hat{\mathcal{I}}_\texttt{G}(F_{ j \to i}, F^\prime_i;\theta_\texttt{G})$ and $\mathcal{L}_\texttt{LMI} = -\frac{1}{N}\sum^N_{j=1}\hat{\mathcal{I}}_\texttt{L}(F_{ j \to i}, F^\prime_i;\theta_\texttt{L})$. By this definition, $\mathcal{L}_\texttt{MI}$ actually corresponds to $\mathcal{L}_\texttt{CTRS}$. The pseudocode for training CMiMC is presented in Algorithm 1 in Appendix A.3.

\begin{table*}[htb]
    \centering
    \small 
    \begin{tabular}{@{}lcccccc@{}}
    \toprule
    \multirow{2}*{\textbf{Method}} & \multirow{2}*{\textbf{Fusion Mode}} & \multicolumn{2}{c}{\textbf{Average Precision (AP)}} & \multicolumn{2}{c}{\textbf{Noisy Case (AP@0.5)}} &\textbf{Communication} \\ &  &  IoU $=0.5$ & IoU $=0.7$ & std $=0.2$m & std $=0.4$m &\textbf{Volume} (KB)\\
    \midrule
    No Collaboration & \usym{2613} & 45.82 & 41.89 & 45.82 & 45.82 & $0.00\times 10^{0}$\\
    Late collaboration & Late & 48.40 & 43.23 & - & - & $4.30\times 10^{-1}$\\
    \midrule
    Who2com (ICRA 2020) & Interm. & 46.31 & 42.02 & 46.19 & 46.03 & $1.05\times 10^3$\\
    When2com (CVPR 2020) & Interm. & 46.71 & 42.42 & 46.64 & 46.18 & $1.05\times 10^3$\\
    V2VNet (ECCV 2020) & Interm. & 59.08 & 52.50 & 57.69 & 54.49 & $2.99\times 10^3$\\
    DiscoNet w/o KD (NeurIPS 2021) & Interm. & 58.91 & 53.33 & - & - & $1.05\times 10^3$\\
    DiscoNet (NeurIPS 2021) & Interm. & 59.74 & 53.43 & 57.71 & 57.38 & $1.05\times 10^3$\\
    V2X-ViT (ECCV 2022) & Interm. & 57.30 & 52.16 & 56.64 & 53.92 & $1.05\times 10^3$\\
Where2comm (NeurIPS 2022) & Interm. & 59.10 & 52.20 & 57.92 & 57.52 &$1.05\times 10^3$\\
    \textbf{CMiMC} & \textbf{Interm.} & \textbf{61.58} & \textbf{55.80} & \textbf{58.80} & \textbf{58.02} & $1.05\times 10^3$\\
    CMiMC ($\sfrac{1}{32}$ compression ratio) & Interm. & 59.78 & 54.04 & 58.94 & 57.52 & $\mathbf{3.28\times 10^1}$\\
    \midrule
    Early Collaboration  & Early & 63.29 & 60.20 & - & - & $3.25\times 10^3$\\
    \bottomrule
    \end{tabular}
    \caption{Performance comparison on V2X-Sim 1.0.}
    \label{tab_APs}
\end{table*}

\section{Experiments and Discussions}
\label{sec:experiments}
\paragraph{Experimental settings.}
We evaluate CMiMC on the V2X-Sim 1.0 dataset~\cite{li2021learning}. V2X-Sim 1.0 is a multi-agent 3D object detection dataset created through the joint simulation of CARLA-SUMO~\citep{dosovitskiy2017carla}. 
It consists of $10^4$ frames from 100 scenes, and each scene contains 2-5 vehicles.  
We cropped the original point cloud to a region defined by $[-32,32]\times[-32,32]\times[-3,2]$m (meters) and discretized it into a BEV map of size $(256,256,13)$. Each voxel has dimensions of 0.25m in length and width and 0.4m in height. 
The architectures of associated encoders/decoders are given in Appendix A.1. The batch size is set to 4. The initial learning rate for CMiMNet is $10^{-4}$, and that for other components is $10^{-3}$, both adopting a multi-step learning rate decay strategy. 
All experiments were conducted using PyTorch on two Nvidia Ampere A40 GPUs.

\paragraph{Benchmarks.}
The experiments use 10 benchmarks: No Collaboration (i.e., single-agent perception), Early Collaboration, Late Collaboration, Who2com~\citep{liu2020who2com}, When2com~\citep{liu2020when2com}, V2VNet~\citep{wang2020v2vnet}, Where2comm \cite{hu2022where2comm}, DiscoNet~\citep{li2021learning}, V2X-ViT~\citep{xu2022v2x} and a variant of DiscoNet w/o knowledge distillation (KD). 
\paragraph{Performance comparison with benchmarks.}
Table \ref{tab_APs} compares APs achieved by CMiMC and other benchmarks. We see that Early Collab. and Late Collab. outperform No Collab., with AP improvements of 38.13\%$\backslash$5.63\% and 43.71\%$\backslash$3.20\% at IoU $= 0.5\backslash0.7$. All Intermediate Collab. methods outperform No Collab., and CMiMC achieves the highest AP 61.58\% (at IoU $= 0.5$), which is close to Early Collab. CMiMC improves the SOTA performance of DiscoNet by 3.08\%$\backslash$4.44\% at IoU 0.5$\backslash$0.7. DiscoNet applies knowledge distillation (KD), which uses a Early Collab. teacher to guide the training of collaboration encoder. Therefore, the performance of DiscoNet depends on the quality of Early Collab. This becomes a bottleneck for DiscoNet. Comparing DiscoNet w/o KD and DiscoNet, we see that using KD only improves AP by 1.41\%$\backslash$0.19\% at IoU 0.5$\backslash$0.7. By contrast, CMiMC uses contrastive learning to train the collaboration encoder in an unsupervised manner, thereby eliminating the above bottleneck. Compared to DiscoNet w/o KD, CMiMC provides AP improvement by 4.53\%$\backslash$4.63\%, surpassing that of KD in DiscoNet. 

\begin{figure}[tb]
\centering
\subcaptionbox{Trade-off in AP@ IoU 0.5}
    {
    \includegraphics[width=0.22\textwidth]{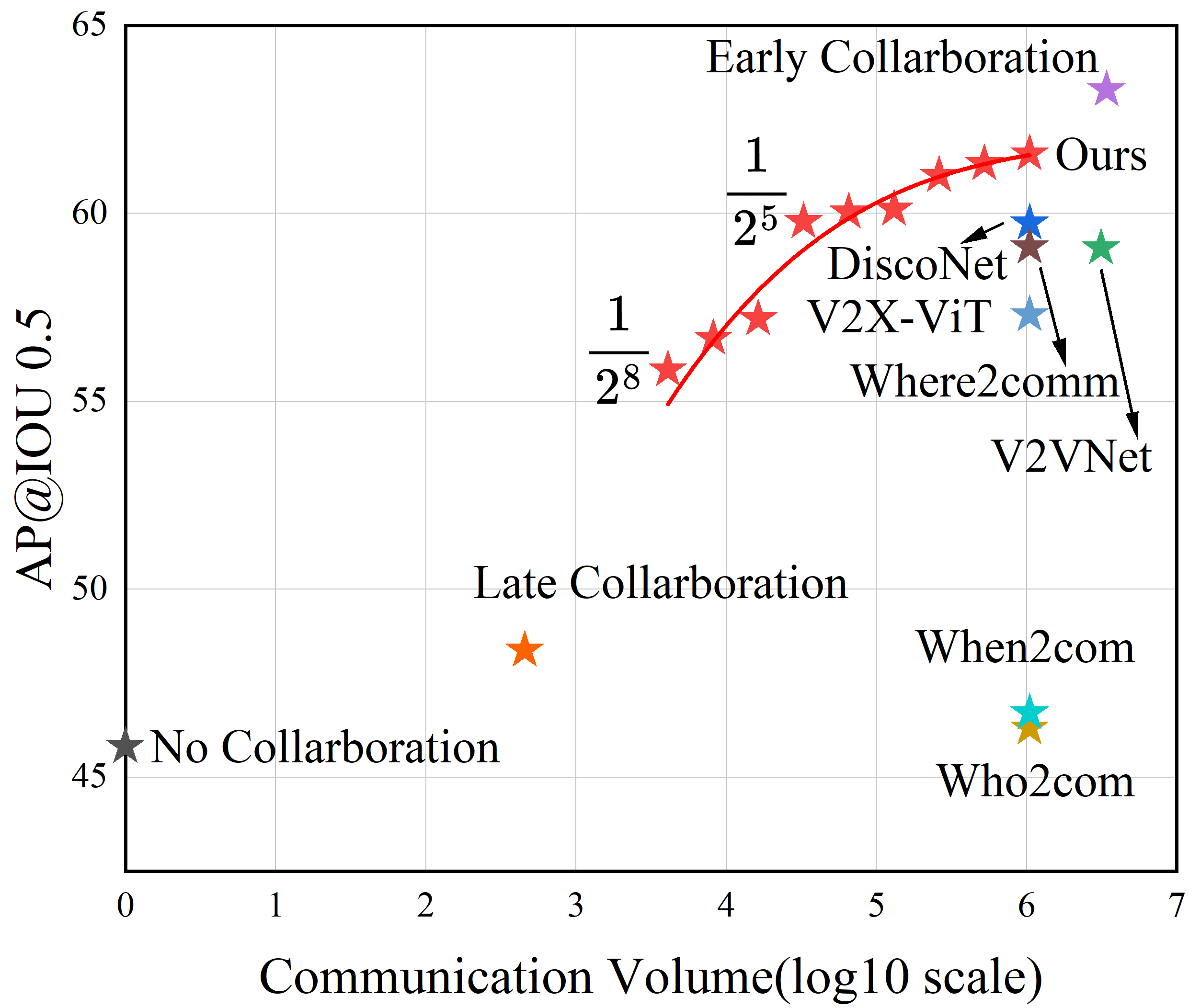}
    }
\subcaptionbox{trade-off in AP@IoU 0.7}
    {
    \includegraphics[width=0.22\textwidth]{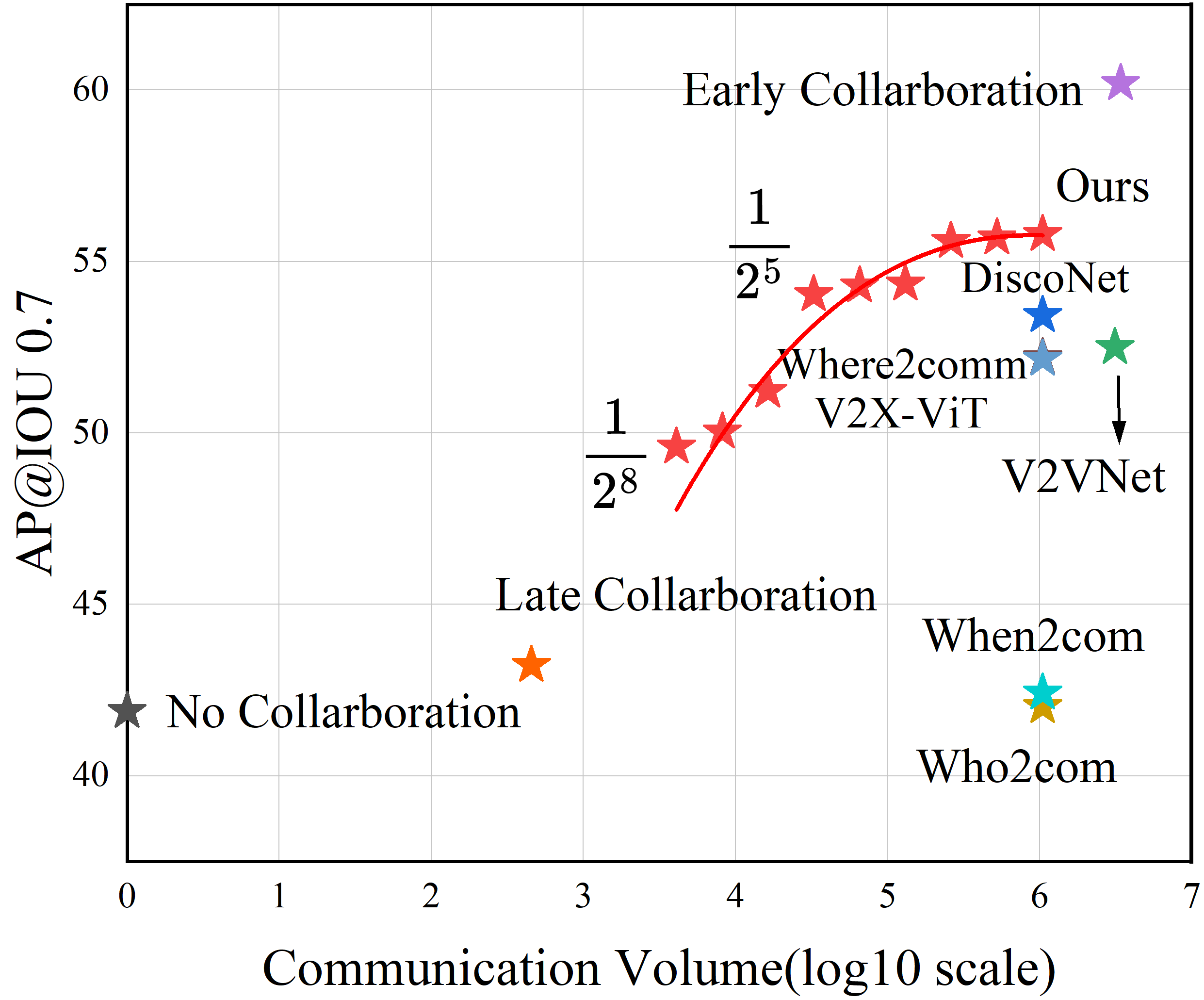}
    }
\caption{Perfomance-bandwidth trade-off of CMiMC.}
\label{fig_perfomance_bandwidth}
\end{figure}
\begin{figure}[tb]
\centering
\subcaptionbox{No Collab. \label{subfig_no}}
    {
    \includegraphics[width=0.14\textwidth]{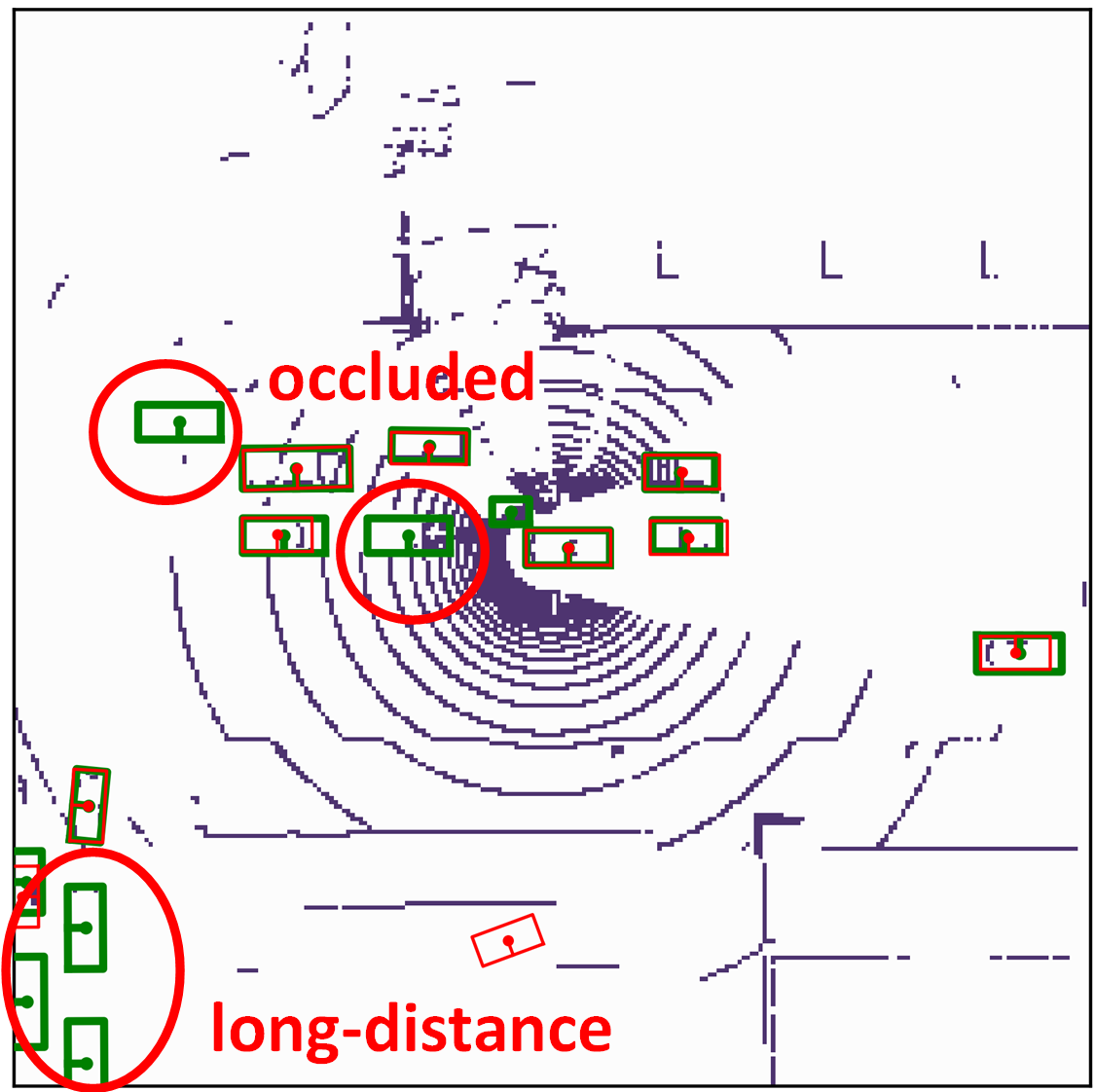}
    }
\subcaptionbox{CMiMC (ours)\label{subfig_ours}}
    {
    \includegraphics[width=0.141\textwidth]{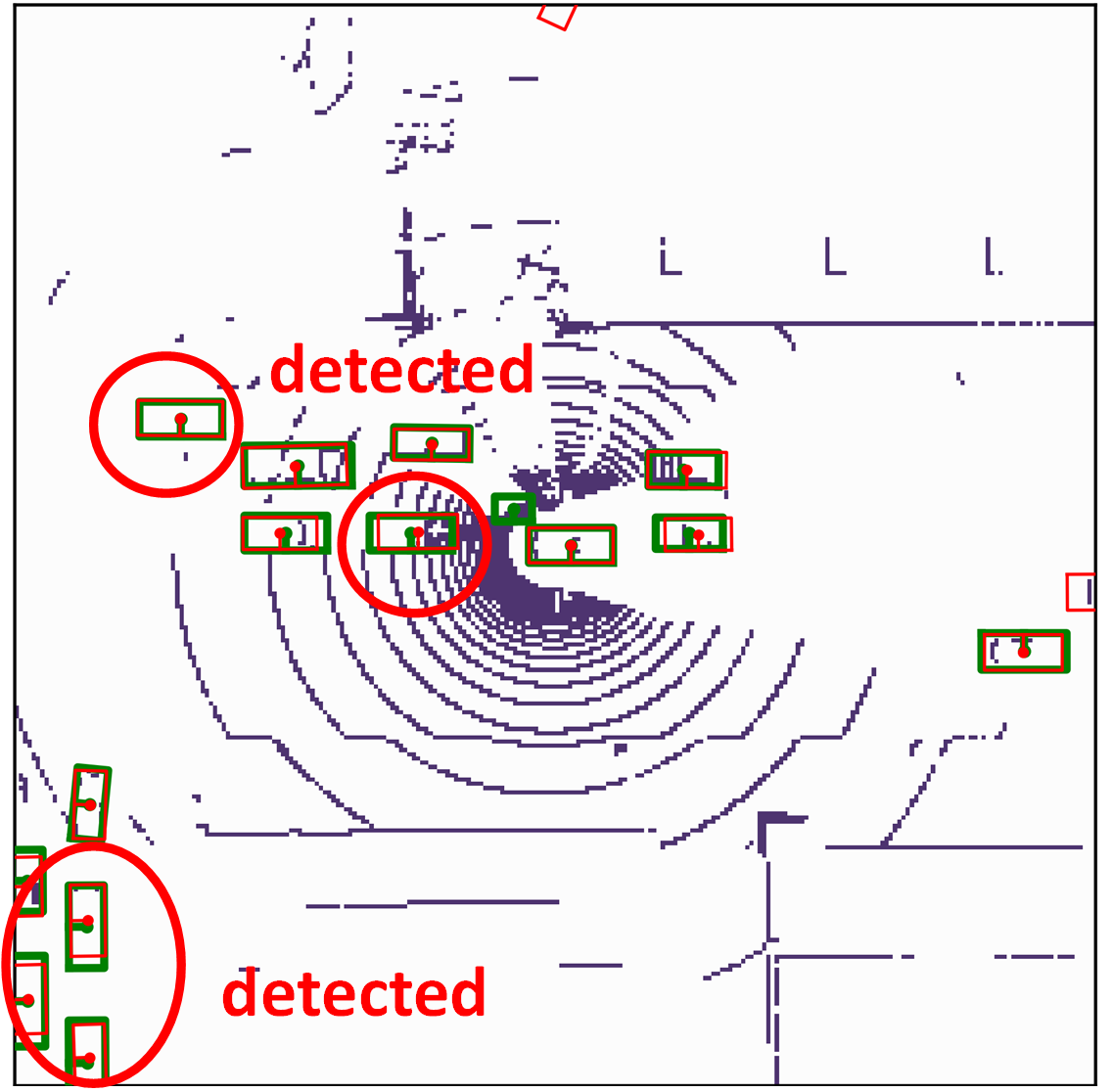}
    }
\subcaptionbox{Early Collab. \label{subfig_early}}
    {
    \includegraphics[width=0.14\textwidth]{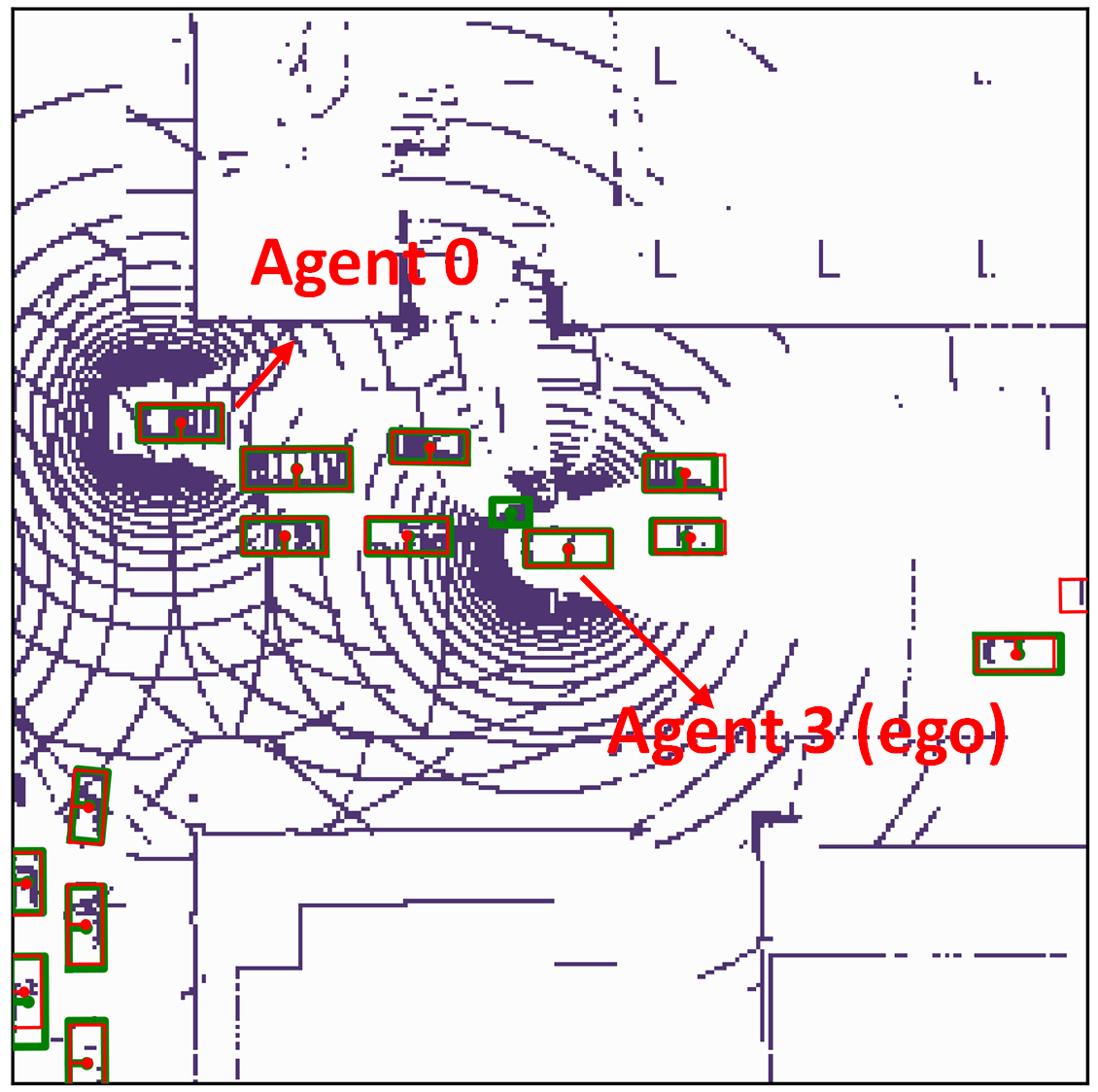}
    }
\caption{No Collaboration v.s. CMiMC v.s. Early Collaboration. Green/Red boxes are ground-truth/predicted boxes.}
\label{fig_det_results}
\end{figure}

\begin{figure*}[t]
\centering
\subcaptionbox{Early Collab.\label{subfig_early_attention}}
    {
    \includegraphics[width=0.145\textwidth]{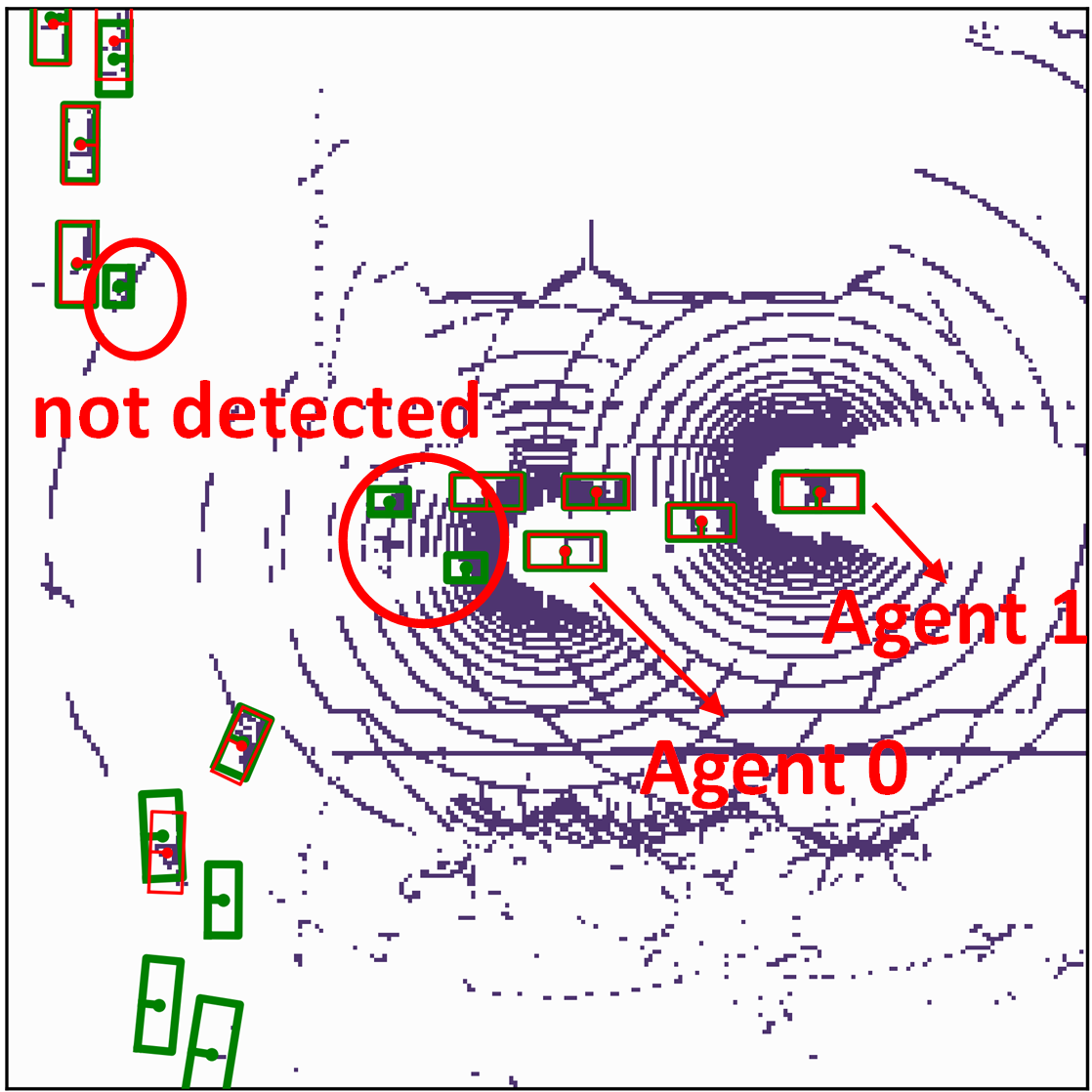}
    }
\subcaptionbox{When2com}
    {
    \includegraphics[width=0.145\textwidth]{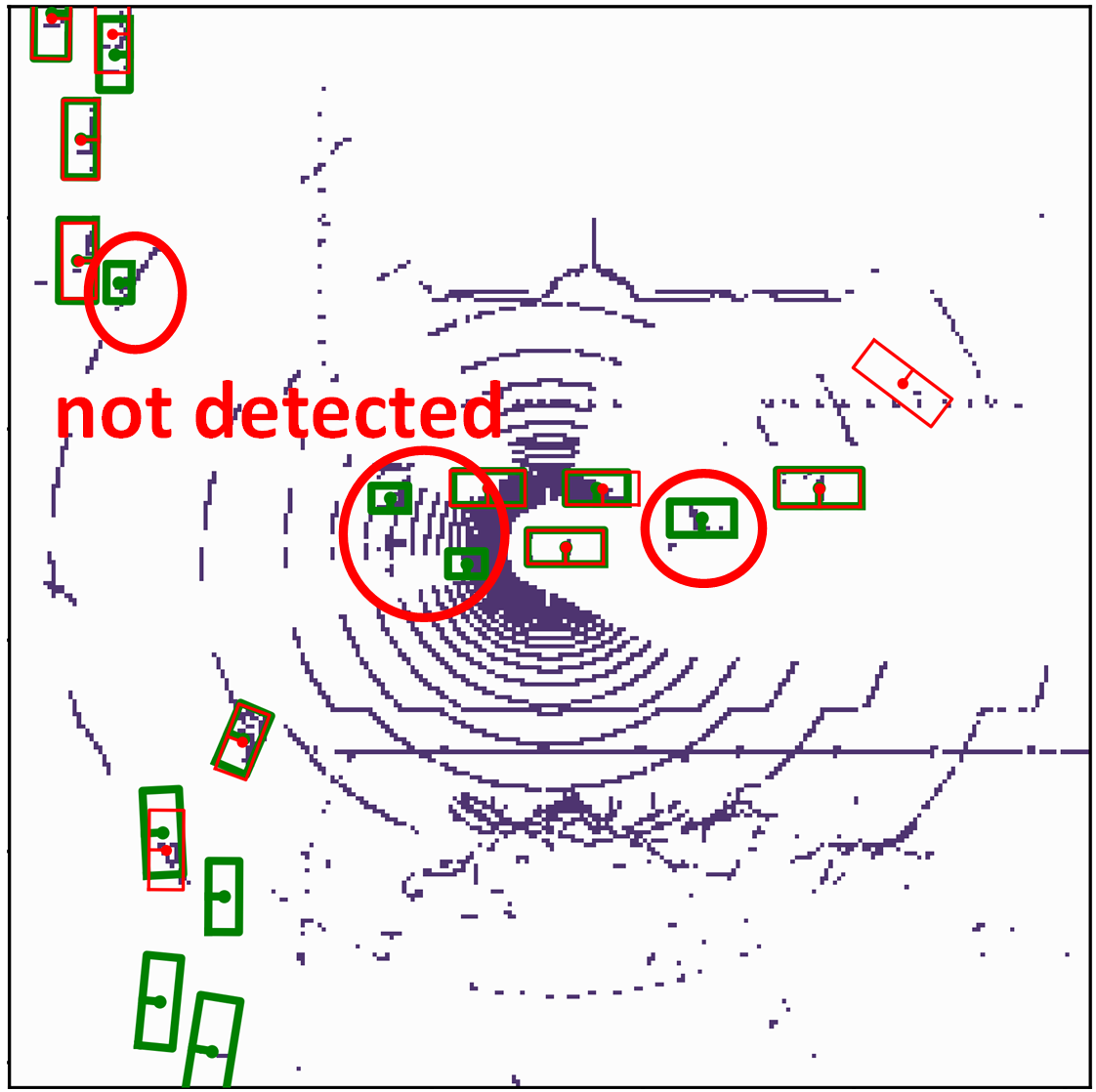}
    }
\subcaptionbox{DiscoNet\label{subfig_disco_attention}}
    {
    \includegraphics[width=0.145\textwidth]{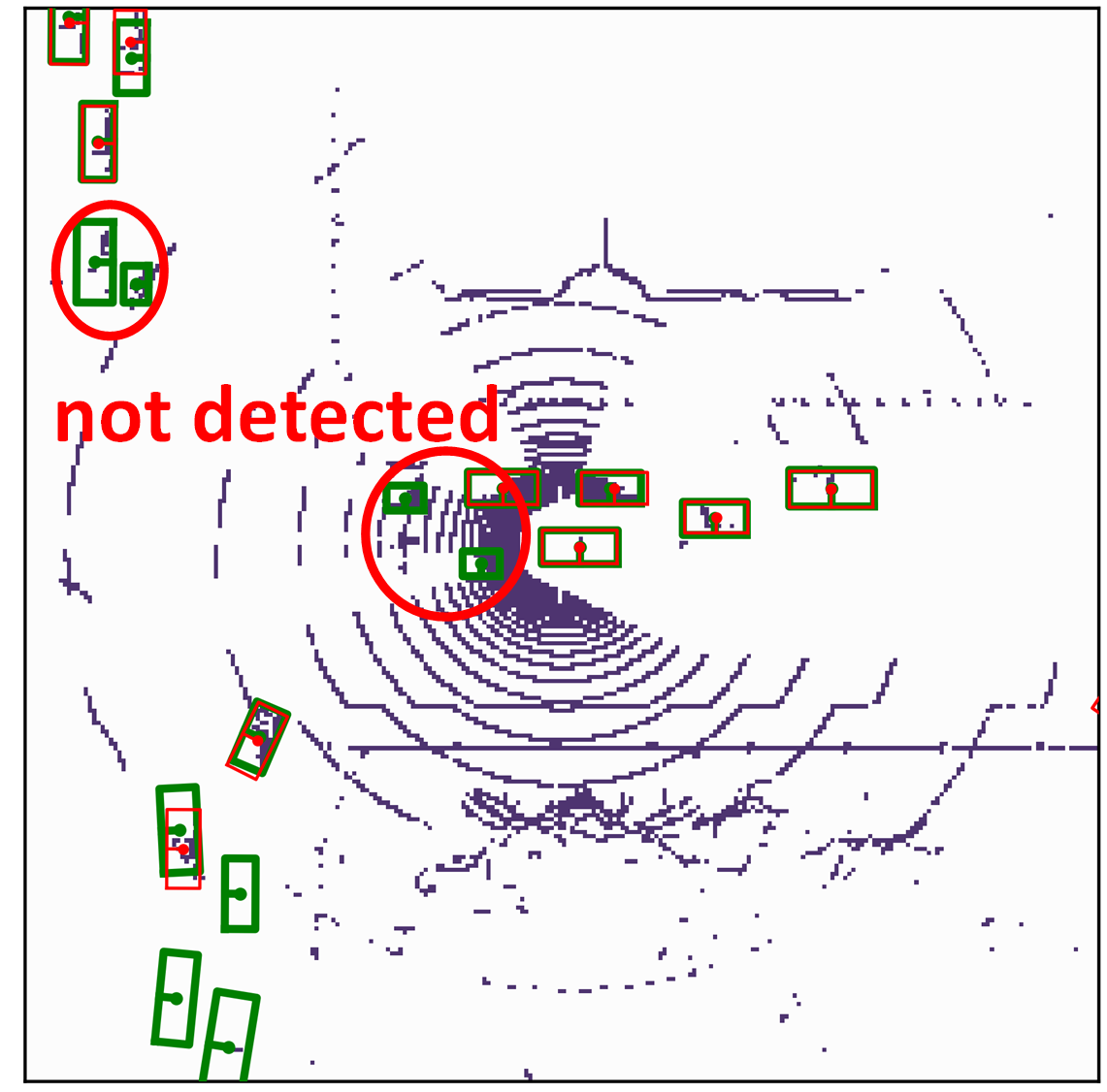}
    }
\subcaptionbox{CMiMC (ours)\label{subfig_ours_attention}}
    {
    \includegraphics[width=0.145\textwidth]{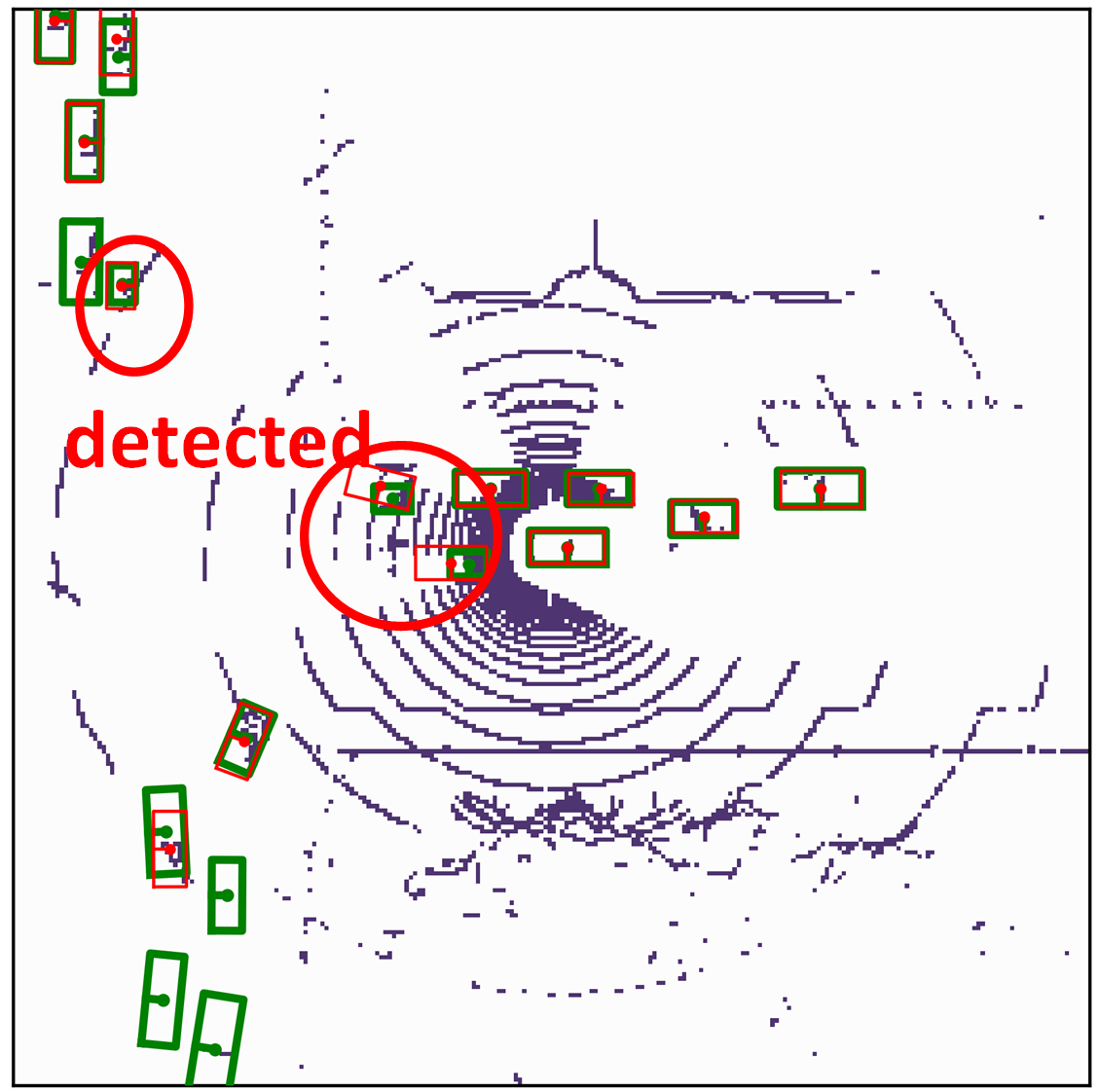}
    }
\subcaptionbox{$W_{0\rightarrow0}$ (CMiMC)\label{subfig_ours_attention_w00}}
    {
    \includegraphics[width=0.15\textwidth]{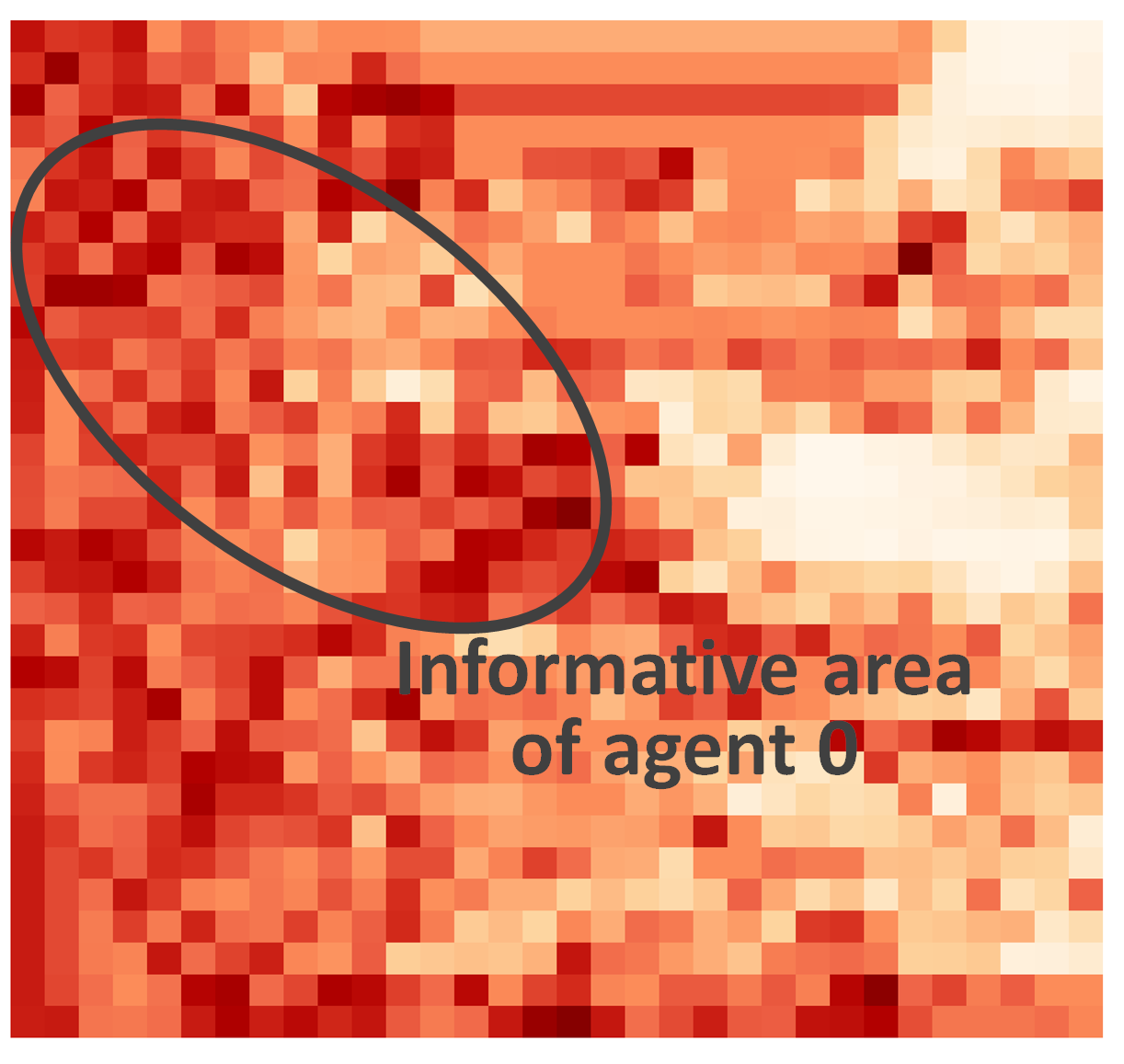}
    }
\subcaptionbox{$W_{1\rightarrow0}$ (CMiMC)\label{subfig_ours_attention_w10}}
    {
    \includegraphics[width=0.175\textwidth]{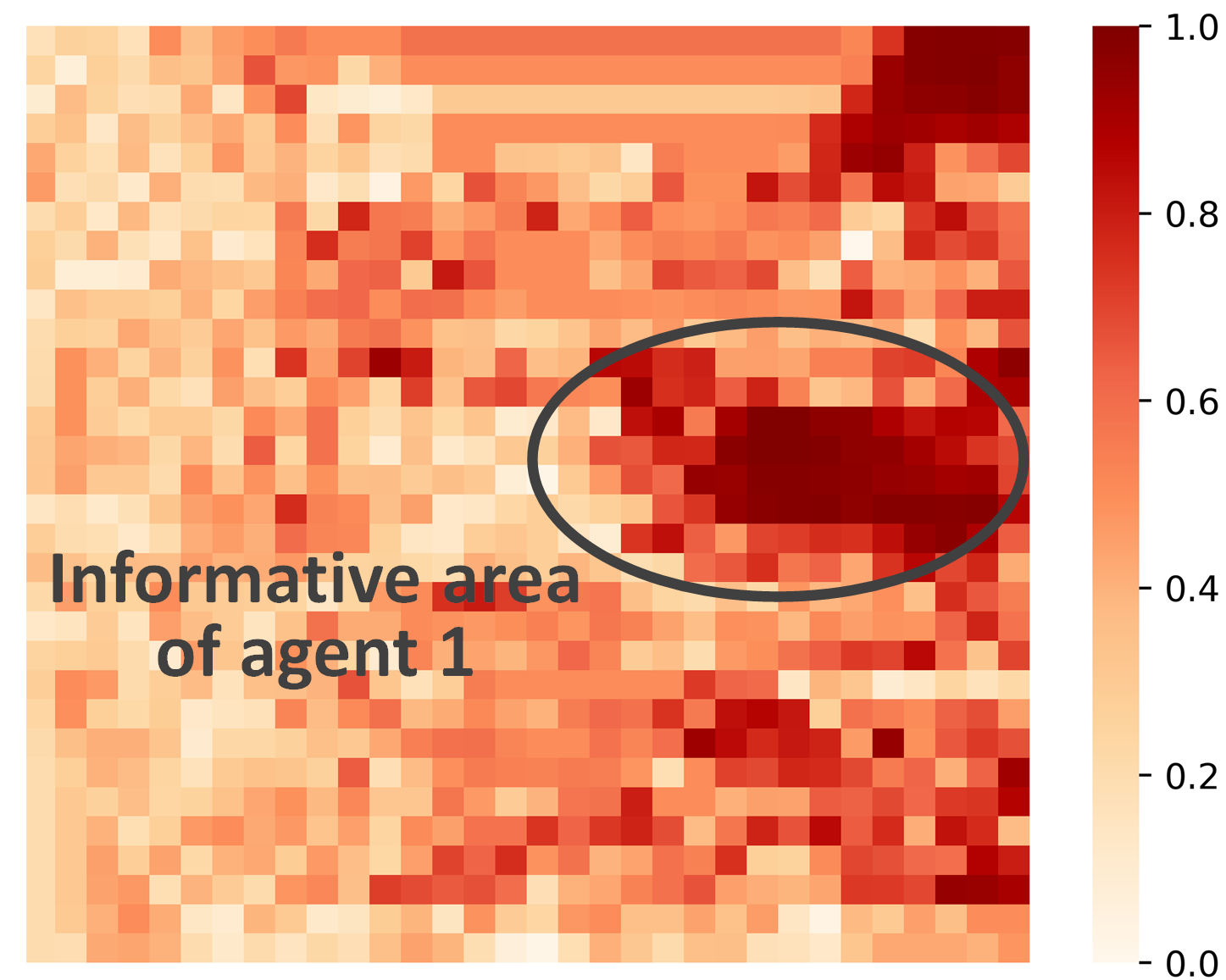}
    }
\caption{Detection results of Intermediate Collab. methods (at agent 0). 
The feature weight matrices are visualized as heatmaps, where darker colors indicate higher weight in the corresponding regions and lighter colors indicate lower weight.}
\label{fig_det_results_attention}
\end{figure*}

\begin{figure*}[t]
\centering
\subcaptionbox{Detection results\label{subfig_weights_det}}
    {
    \includegraphics[width=0.165\textwidth]{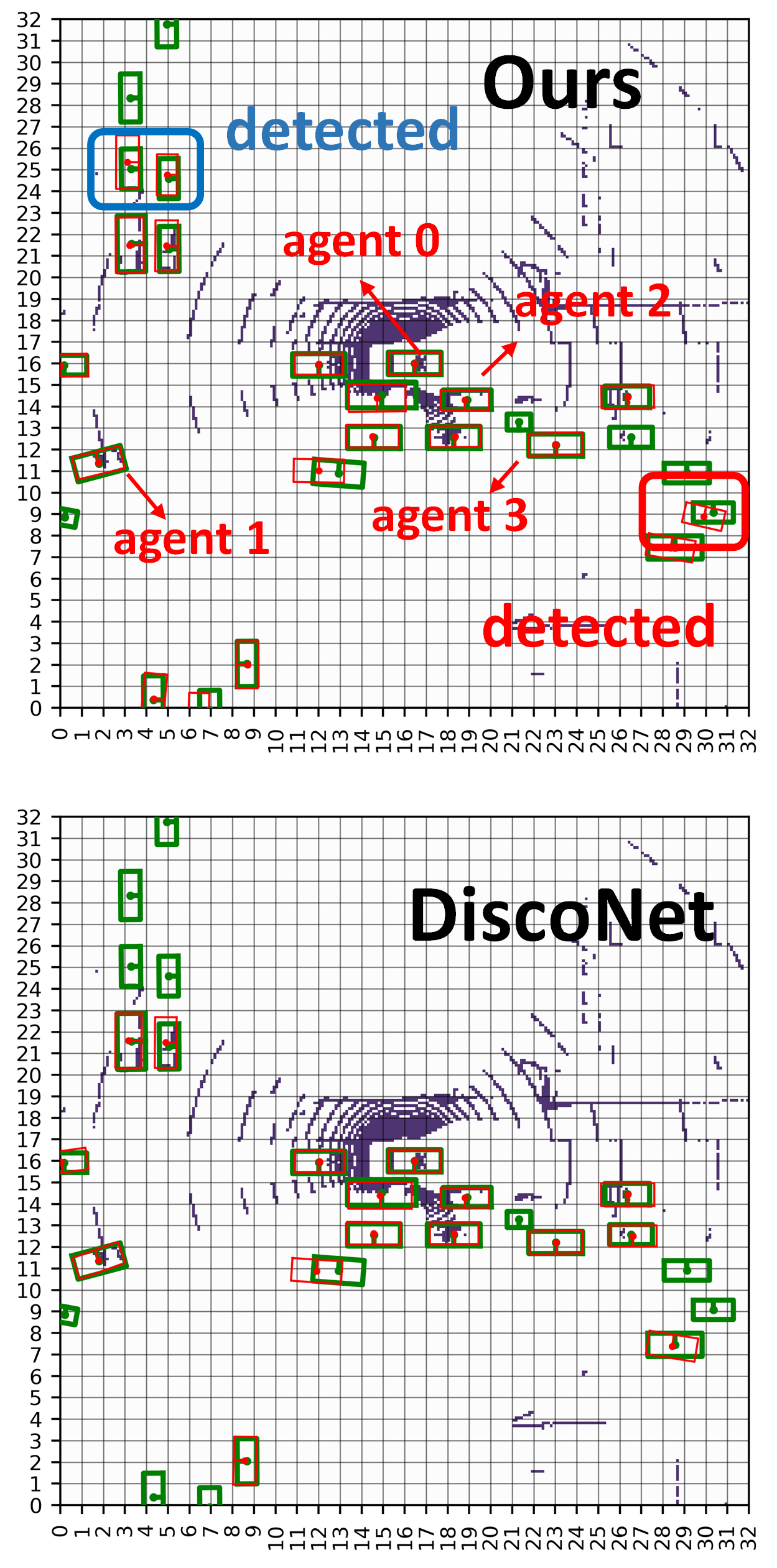}
    }
\subcaptionbox{$W_{0\rightarrow0}$\label{subfig_weights_w00}}
    {
    \includegraphics[width=0.173\textwidth]{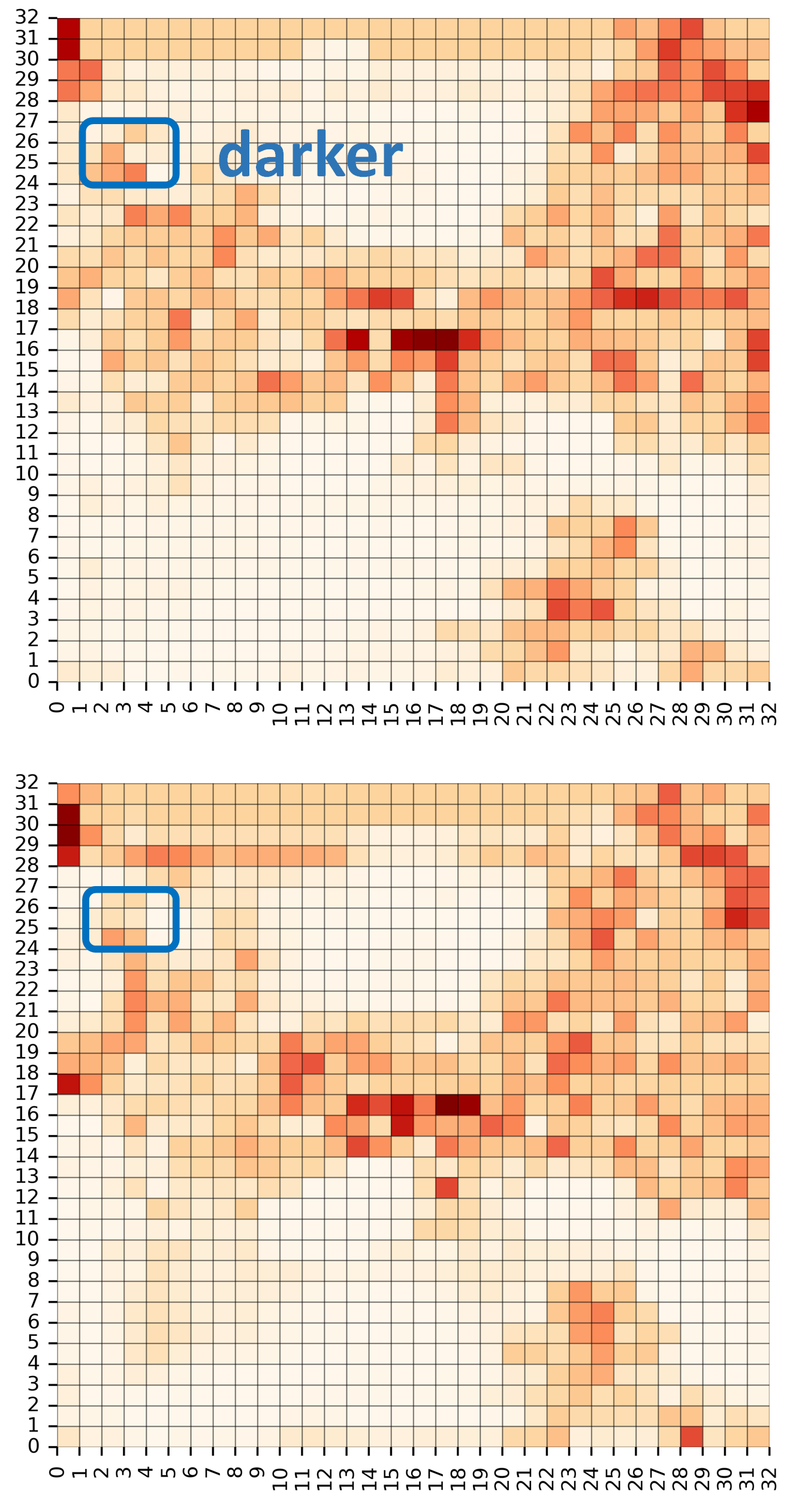}
    }
\subcaptionbox{$W_{1\rightarrow0}$\label{subfig_weights_w10}}
    {
    \includegraphics[width=0.171\textwidth]{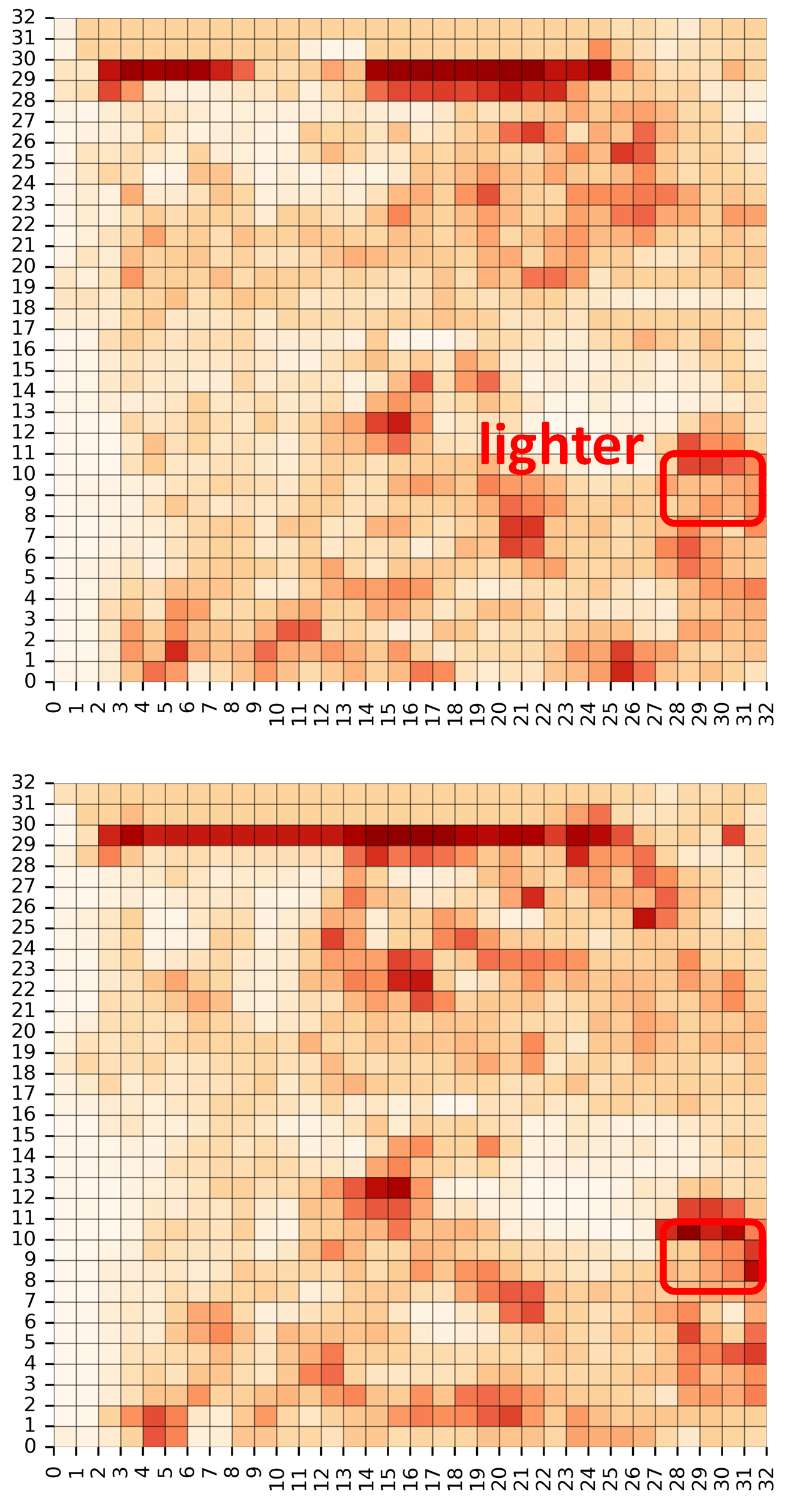}
    }
\subcaptionbox{$W_{2\rightarrow0}$\label{subfig_weights_w20}}
    {
    \includegraphics[width=0.17\textwidth]{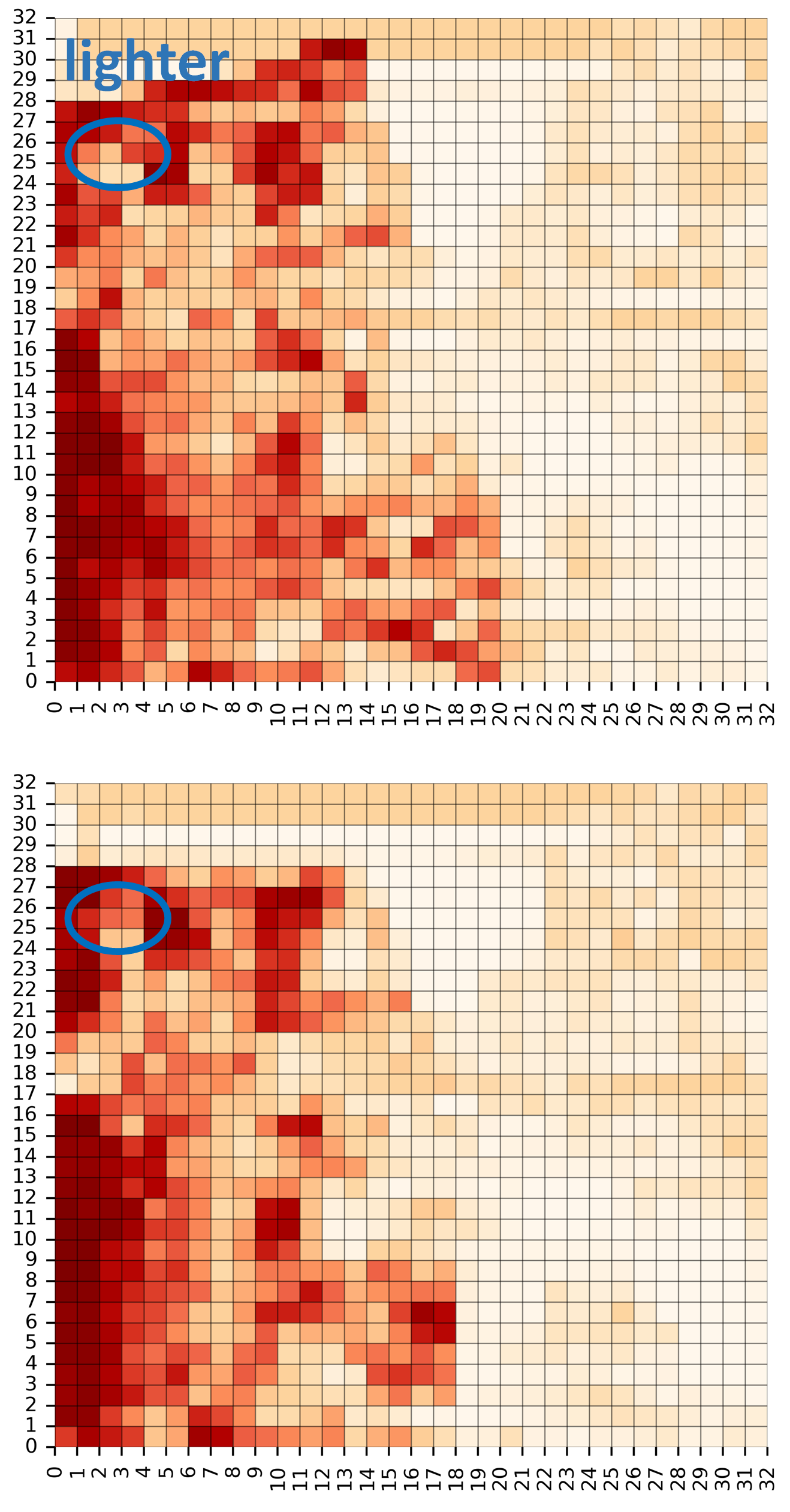}
    }
\subcaptionbox{$W_{3\rightarrow0}$\label{subfig_weights_w30}}
    {
    \includegraphics[width=0.199\textwidth]{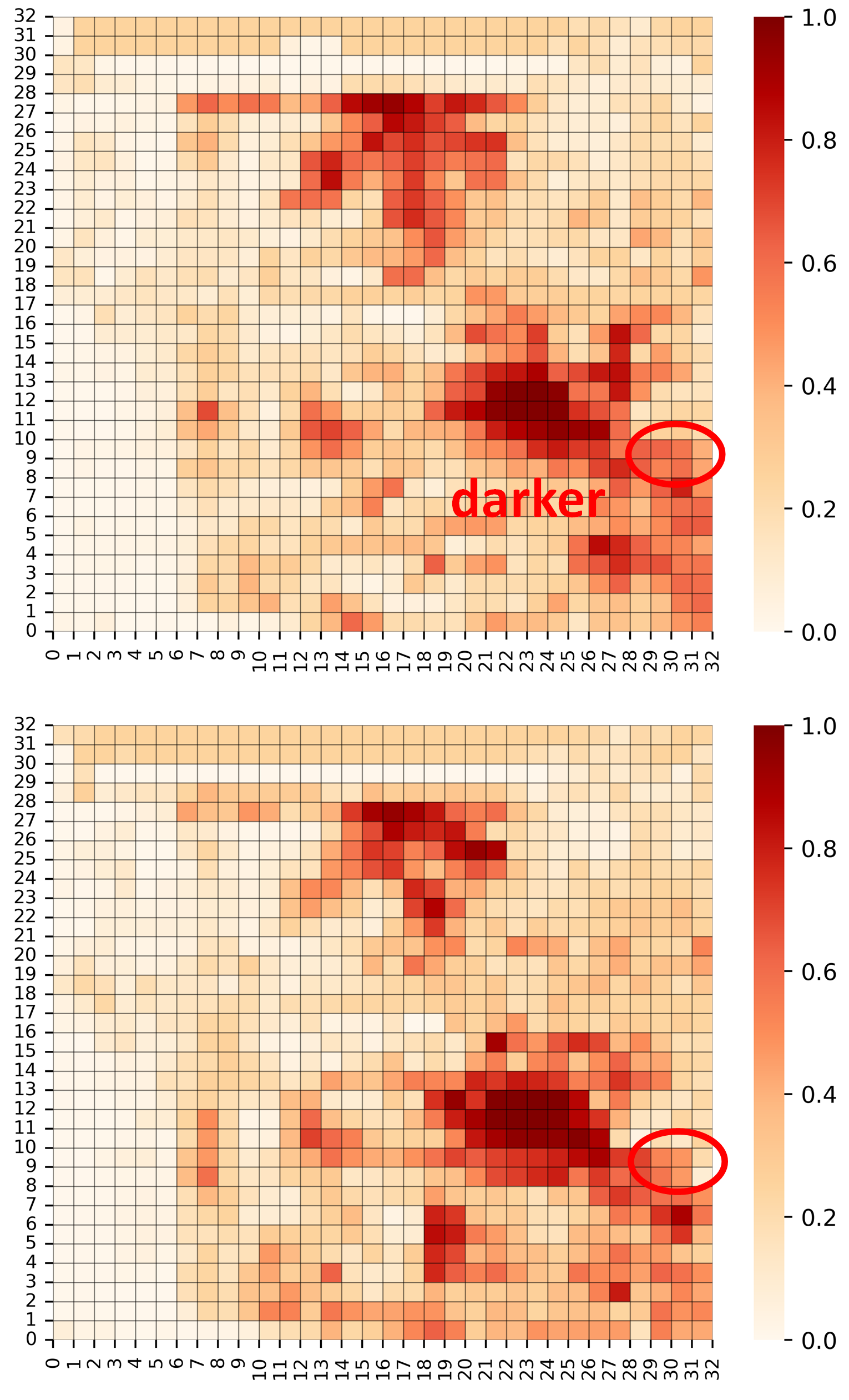}
    }
\caption{Detection results and feature weight matrix of CMiMC and DiscoNet (at agent 0).}
\label{fig_weights}
\end{figure*}

\paragraph{Robustness to localization noise.}
We also consider the case of localization noise where the random noise (with a mean of 0m and a standard deviation of $0\backslash0.2\backslash0.4$m) is added to the pose information of agents. Based on the results in Table \ref{tab_APs}, we see a general trend that the performance degrades with the increase in localization noise. However, the proposed CMiMC still outperforms other benchmarks in all examined cases of localization noise.

\paragraph{Perfomance-bandwidth trade-off.}
Table \ref{tab_APs} also shows the communication volume of CMiMC and other benchmarks. No Collab. uses single-agent perception and hence does not incur communication overhead. While Early Collab. achieves the highest AP, it also requires the largest communication volume ($3.13 \times 10^3$KB) due to transmitting raw data of LiDAR point clouds. Late Collab. has a low communication volume but also has low AP. Intermediate collab. methods strike a balance between performance and bandwidth usage. Notably, our CMiMC provides higher AP than other Intermediate Collab. methods with comparable or even lower communication volume.

Fig.~\ref{fig_perfomance_bandwidth} further shows the performance-bandwidth trade-off of CMiMC. We employ compression techniques (details of compression technique are given in Appendix A.4) to reduce the size of intermediate features. We use 8 compress ratios $\sfrac{1}{2^n}, n=1,2,\dots,8$ and show the corresponding AP achieved by CMiMC. It can be observed that with compression ratio $\sfrac{1}{32}$, CMiMC still achieves a higher AP than the SOTA performance of DiscoNet. Particularly, even with the maximum compression ratio, i.e., $\sfrac{1}{256}$, CMiMC still outperforms When2com, Who2com, and Late Collab.

\paragraph{Visualization of detection results.}
Fig.~\ref{fig_det_results} shows detection results generated by CMiMC, No Collab., and Early Collab. 
We see in Fig.~\ref{subfig_no} that No Collab. cannot perceive objects in occluded and long-range areas. Fig.~\ref{subfig_ours} shows that CMiMC can overcome these issues by leveraging Intermediate Collab. Further comparing it with Fig.~\ref{subfig_early}, we see that CMiMC provides perception results similar to Early Collab.

\paragraph{Attention mechanisms.}
Fig.~\ref{fig_det_results_attention} shows results of methods that involve attention mechanisms (i.e., CMiMC, When2com, and DiscoNet). We see that CMiMC detects more objects than When2com and DiscoNet. The performance inferiority of When2com is due to the utilization of scalar attention which cannot accurately capture the regional characteristics in views. DiscoNet also utilizes voxel-level attention weights, but its performance is bottlenecked by the Early Collab. teacher. As shown in Fig.~\ref{subfig_early_attention} and ~\ref{subfig_disco_attention}, objects not detected by Early Collab. are often missed by DiscoNet as well. In contrast, CMiMC can learn better attention weights (Fig.~\ref{subfig_ours_attention_w00},~\ref{subfig_ours_attention_w10}), which better capture the importance of regional feature values at a voxel-level resolution.

\paragraph{Visualization of feature weight matrices.}
Fig.~\ref{fig_weights} compares the feature weight matrices and perception results of CMiMC and DiscoNet. The results indicate that CMiMC can identify critical regions more efficiently compared to DiscoNet. For example, for the two vehicles in the top left corner (detected successfully by CMiMC but missed by DiscoNet), CMiMC gives higher weight for the view of agent 0 as it captures discriminative information about these vehicles and decreases weight for the view of agent 2 as such information is not contained there.

\paragraph{Supplementary results.} We provide supplementary experimental results in Appendix B, including hyper-parameter selection, time complexity, loss curves, and evaluation on more public datasets (DAIR-V2X~\citep{yu2022dair}, V2X-Sim 2.0~\citep{li2022v2x}, V2XSet~\citep{xu2022v2x} and V2V4Real~\citep{xu2023v2v4real}).

\section{Conclusion}
\label{sec:Conclusion}
This paper proposes an intermediate collaboration method called CMiMC, which presents a novel way of constructing good collaborative views through the maximization of mutual information. The core of CMiMC is CMiMNet which estimates and maximizes the multi-view mutual information between a collaborative view and multiple individual views. CMiMC enables the collaboration encoder to efficiently identify beneficial regional information in individual views and aggregate them properly into the collaborative view. Comprehensive experimental results demonstrate that CMiMC achieves an excellent trade-off between perception performance and communication bandwidth, and outperforms SOTA strategies. CMiMC is a generic framework and can be applied to various application scenarios, e.g., autonomous driving, robotics, and surveillance.


\section*{Acknowledgments}
The work of Wanfang Su and Lixing Chen was partially supported by the National Natural Science Foundation of China (NSFC) under Grant 62202293 and 62372297. The work of Yang Bai was partially supported by NSFC under Grant 62303306. The work of Zhe Qu was partially supported by NSFC under Grant 62302525.

\bibliography{aaai24}
\end{document}